\documentclass[preprint,12pt]{elsarticle}

\usepackage{amssymb}

\usepackage{subcaption}
\usepackage{stix}
\usepackage{xcolor} 
\definecolor{burgundy}{rgb}{0.5, 0.0, 0.13}
\usepackage[parfill]{parskip} 
\usepackage{pdflscape}
\usepackage{url}
\usepackage{setspace}
\usepackage{lineno}
\usepackage{multirow}

\journal{Computers and Electronics in Agriculture}

\begin{document}

\begin{frontmatter}



\title{Lincoln's Annotated Spatio-Temporal Strawberry Dataset (LAST-Straw)}


\author[inst1,inst2]{K.M.F. James$^1$}
\author[inst1]{K. Heiwolt$^1$}
\author[inst2]{D.J. Sargent}
\author[inst1]{G. Cielniak}

\affiliation[inst1]{organization={University of Lincoln},
            addressline={Brayford Pool}, 
            city={Lincoln},
            postcode={LN6 7TS},            
            country={United Kingdom}}

\affiliation[inst2]{organization={NIAB East Malling},
            addressline={New Road}, 
            city={East Malling},
            postcode={ME19 6BJ}, 
            state={Kent},
            country={United Kingdom}}

\begin{abstract}
    Automated phenotyping of plants for breeding and plant studies promises to provide quantitative metrics on plant traits at a previously unattainable observation frequency. Developers of tools for performing high-throughput phenotyping are, however, constrained by the availability of relevant datasets on which to perform validation. To this end, we present a spatio-temporal dataset of 3D point clouds of strawberry plants for two varieties, totalling 84 individual point clouds.
    
    We focus on the end use of such tools - the extraction of biologically relevant phenotypes - and demonstrate a phenotyping pipeline on the dataset.  This comprises of the steps, including; segmentation, skeletonisation and tracking, and we detail how each stage facilitates the extraction of different phenotypes or provision of data insights. 
    
    We particularly note that assessment is focused on the validation of phenotypes, extracted from the representations acquired at each step of the pipeline, rather than singularly focusing on assessing the representation itself. Therefore, where possible, we provide \textit{in silico} ground truth baselines for the phenotypes extracted at each step and introduce methodology for the quantitative assessment of skeletonisation and the length trait extracted thereof.
    
    This dataset contributes to the corpus of freely available agricultural/horticultural spatio-temporal data for the development of next-generation phenotyping tools, increasing the number of plant varieties available for research in this field and providing a basis for genuine comparison of new phenotyping methodology. 
\end{abstract}

\begin{graphicalabstract}
\includegraphics[width=\textwidth]{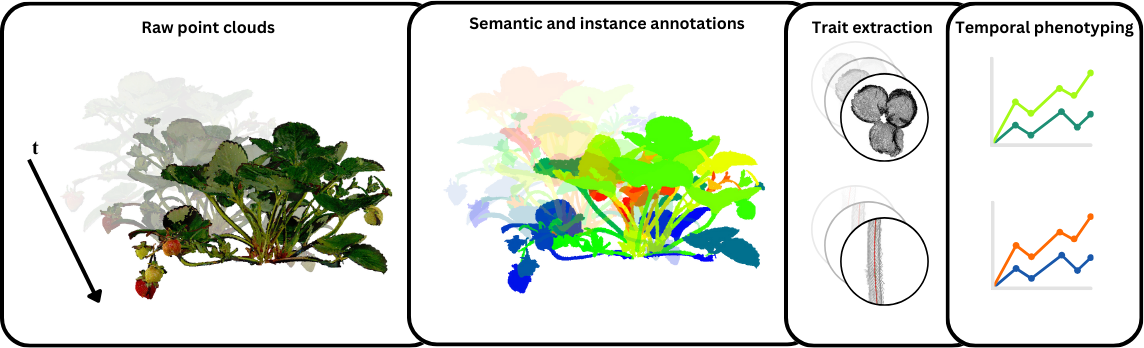}
\end{graphicalabstract}

\begin{highlights}
\item A 3D spatio-temporal dataset of 84 strawberry plant scans, spanning developmental stages from seedling to mature.
\item Prerequisite annotation steps (semantic, instance and stem skeletons) enabling trait extraction and methodology for evaluating the quality of these.
\item Demonstrated phenotyping pipeline and extracted baselines at individual time steps and illustrated temporal data insights.
\end{highlights}


\end{frontmatter}

\section{Introduction}



Plant breeders aim to screen all progeny within a breeding population at the appropriate moment to observe all the favourable characteristics of any given individual and make a selection decision based on those observed phenotypes. The challenge with this process in strawberry variety development is that breeding populations can comprise up to 20,000 genetically unique individuals and there are a range of complex, agronomically important phenotypes including flowering time, structure of floral organs, complexity of fruit trusses, length of peduncles, display of fruit, sweetness, fruit colour, shape and size, canopy architecture and production of runners to name but a few traits. Observations must be made repeatedly in order to fully evaluate all essential characteristics, and the window of opportunity when plants are fruiting is short, meaning that traditional qualitative selection by breeders often does not capture all the potentially observable variation within a breeding population~\cite{mathey2013}. 

Developments in computer vision and sensing have led to the emergence of a growing presence of high-throughput phenotyping tools \cite{danilevicz2021_review,li2020_review}. More recently, developments in three-dimensional (3D) perception have shown additional promise for phenotyping - allowing for greater exploration of the geometry of the plant or its organs \cite{paulus2019_review}. This is particularly important for performing measurements, where 3D data has been shown to substantially improve the accuracy when measuring structures within a plant \cite{boogaard2023}.

Increasing levels of automation also enable higher observation frequencies and the ability to associate and integrate observations across time digitally is instrumental in capturing the substantial temporal variations in plant morphology. In \cite{das2019leveraging}, the authors identify a pressing need for such whole-of-lifecycle phenotyping and call for the development and study of novel temporal phenotypes, i.e. traits that are by definition dynamic, such as the emergence timing of new organs or growth rates. 

A ubiquitous challenge in research centring on the development of machine vision methods for agriculture is access to relevant datasets \cite{ jiang2020_review}. The collection and annotation of such datasets is time-consuming and resource-intensive in terms of man-hours. Thus, the free release and sharing of 3D point cloud datasets is essential to facilitate the development of next-generation phenotyping tools \cite{danilevicz2021_review}. However, 3D spatio-temporal datasets of crops are limited to a few species (see Sec.~\ref{sec:related}). 

It should be remembered that the ultimate users of phenotyping tools are plant breeders selecting particular combinations of traits of interest to meet the demands of propagators, growers, retailers and consumers. It is thus important that the developers of automated phenotyping tools consider this during development. A growing body of research focuses on the development of such tools for phenotyping, although evaluation of the resulting phenotypes against a ground truth is rare (e.g. \cite{he2017}), likely due to the large amount of effort required to measure or annotate ground truth data. A gap also exists between traditional and digital phenotypes, with the latter facilitating the reporting of new or modified phenotypes, which are quantitative descriptors \cite{das_choudhury2020} or capable of quantifying more abstract traits \cite{feldmann2021_review}. Due to the potential for automated phenotyping to enable more frequent, comprehensive, quantitative observations of entire breeding populations, the opportunity to capture all the genetic variation contained within those populations is great and could potentially significantly increase the selection precision of the breeding process. 

Strawberries are a speciality berry crop of major economic importance \cite{hancock2020}, yet research into automated phenotyping of strawberry is relatively unexplored. It has thus far been largely limited to traits relating to fruit and phenology, and other economically important traits such as the measurement of peduncle length have not been addressed \cite{james2022}.

The main contribution of this paper is a 3D spatio-temporal dataset of individual strawberry plants of two varieties (6 plants, 14 time steps), which we envision to be of use to developers and researchers of automated phenotyping tools. This is provided along with manually annotated semantic and instance annotations for 13 plants, and manually tuned `\textit{in silico}' ground truth skeletons for the stems for these samples (totalling 267). The contributed dataset, named LAST-Straw, addresses the lack of available 3D spatio-temporal data for strawberry plants, supplying high-quality data along with rich annotations, to allow for development of automated phenotyping tools. We further demonstrate a phenotyping pipeline (Sec.~\ref{sec:phenotyping}) and provide benchmarks for the phenotypes which can be extracted at each stage, where possible, against ground truth values. The demonstrated traits are organ counts, plant volume, leaf area and stem length. We also demonstrate methodology for assessing skeletonisation results through a phenotyping lens. Finally, in Sec.~\ref{sec:tracking}, we illustrate how tracking instances of organs over time would allow for the tracking of particular phenotypes. Throughout, we highlight challenges that exist in performing phenotyping using real-world data and highlight key areas for future research. 

\section{Related work} \label{sec:related} 

\subsection{Existing 3D datasets at plant level}
Due to the widespread availability and low cost of 2D cameras as well as the extensive literature on well-studied classical computer vision techniques and suitable neural network architectures, the bulk of recent work at the intersection of computer vision and plant phenotyping focuses on the analysis of 2D images. For an overview on 2D imaging datasets, refer to \cite{LU2020Asurvey} - a survey of public datasets for computer vision tasks in precision agriculture.
For the purpose of evaluating plant morphology, however, considerable detail about 3D configurations and areas obscured by occlusion is lost in 2D projections. 

While there has been a notable increase in 3D plant datasets in recent years, the intended application and quality vary widely. 
The majority of publicly available collections are limited by their low spatial resolution. In some instances, datasets can be more accurately classed as 2.5D rather than true 3D data, where depth information is recorded from a single view (usually top-down) or very few perspectives (e.g. 
Purdue Apple Tree~\cite{akbar2016novel}, and 3rd Autonomous Greenhouse Challenge~\cite{petropoulou2023lettuce} datasets). These representations typically lack the complete morphological information required for most 3D analysis. A number of 2.5D datasets also offer time-series measurements for the same plants (e.g. ETH Eschikon Plant Stress Phenotyping Dataset \cite{khanna2019spatio}, 
and MSU-PID \cite{cruz2016multi}), though the reliability of geometric measures is compromised by substantial occlusions inherent to the 2.5D representation.
Other data collections are created with a single specific study or use-case in mind and lack dense annotations (e.g. LFuji~\cite{gene2020lfuji}). Datasets for 3D plant phenotyping have also been made available as a collection of images from different side views, which can later be combined into a 3D representation. The UNL-3DPPD Dataset \cite{das_choudhury2020} makes 10 RGB side views available for 20 maize and sorghum plants, imaged daily. This dataset, however, is not associated with annotations as the authors used a rule-based level-set method for segmentation into leaf or stem classes.

We are aware of three notable datasets, distinguished by remarkably high spatial resolution and rich annotations. Firstly, the Rose-X dataset \cite{dutagaci2020rose} features eleven 3D models of rosebush plants with varying architectural complexity, along with semantic segmentation labels. The scans were recorded using X-ray tomography and are completely free of self-occlusions. 
Secondly, the Soybean-MVS \cite{sun2023soybean} dataset consists of 102 stereoscopic 3D reconstructions of five Soybean varieties. Repeated measurements of the same individual plants cover all 13 stages of their growth period. 3D semantic annotations are also provided.

Finally, Pheno4D \cite{schunck2021} can be considered the most closely associated related work to our dataset. This set contains 12 repeated scans of 7 maize plants and 20 repeated scans of 7 tomato plants each, totalling 140 point clouds, 77 of which are densely annotated with semantic and temporally consistent instance labels. The authors also provide baselines for semantic and instance segmentation, spatio-temporal point cloud registration, and surface reconstruction, as well as \textit{in silico} measurements of leaf length and area and stem lengths.

\subsection{Spatio-temporal strawberry datasets}
In terms of dataset availability for strawberry plants, this is largely limited to 2D datasets collected for particular applications without temporal consideration. A recent publication has seen the publication of a 2D RGB+Orange Cyan Near-infrared (OCN) dataset for fruit growth tracking \cite{wen2024}. In the study, hourly images were captured in situ, providing around 100 images across the ripening period for each fruit. 

A single example exists of a true 3D dataset for strawberry - a dataset of 100 high-quality point clouds of individual strawberry fruits was captured for automated fruit quality assessment \cite{he2017}. The dataset includes 10 varieties of strawberry, purchased from a local supermarket, from which point clouds were captured using a structure from motion approach. A key focus of this study was the validation of phenotypes (achene number, calyx size, colour, height, length, width and volume) against manually measured ground truths, to determine the success of the automated measurements. There is, however, no temporal component to this dataset, and it is limited to individual fruit.

For strawberry plants, there currently are no 3D time-series datasets available comparative to the samples for maize and tomato provided in Pheno4D \cite{schunck2021}. The dataset we present in this paper attempts to address this gap, to enable further research into automated phenotyping of strawberry. In this way, methods for measuring traits beyond those relating to fruit can also be developed and assessed.

\section{Data acquisition}

\subsection{Dataset overview}

This dataset contains colour point clouds of six strawberry plants of two varieties, \textit{Driscoll's} Katrina (variety A) and \textit{Driscoll's} Zara (variety B), with three individuals of each. Each plant underwent data capture 14 times over the span of 11 weeks (13-05-2022 - 29-07-2022), spanning the development of the individual plants from seedling to fruiting. Scans were not taken at equal time steps, but rather with increased frequency when the plant was undergoing more rapid development.

\begin{figure}
    \centering
    \includegraphics[width=\textwidth]{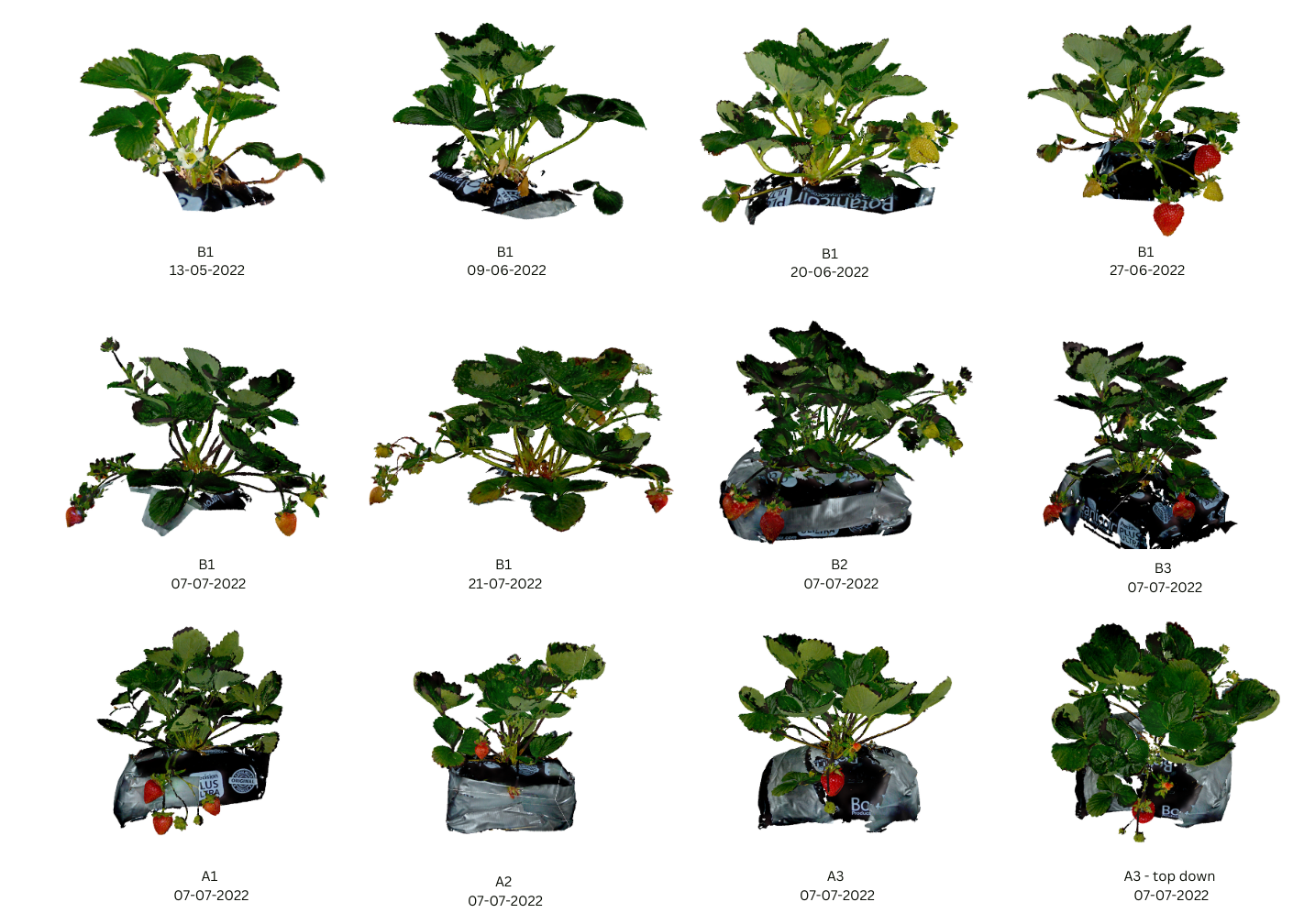}
    \caption{The dataset contains two varieties and captures variation in developmental stages, from young to mature plants. Here we provide 2D renders of a sample of 3D meshes for plant B1, as well as all six plants on 07-07-2022 and a top-down view of plant A3 at this timestep. The main volume of the grow bag was manually removed to aid clarity of visualisation for samples of B1. Note - meshes not to scale, best viewed in colour. }
    \label{fig:dataset_sample}
\end{figure}

A total of 84 point clouds are provided with roughly 0.7 - 1.25M points depending on the complexity of the plant, and an average of 1.17M points per scan. A sample of the dataset is presented in Fig.~\ref{fig:dataset_sample}, in which a sample of the time steps for plant B1 are demonstrated to show variance in maturity, and all six plants for time step 07-07-2022 to illustrate inter- and intra-variety variance across a time step.

\begin{figure}[hbt!]
    \centering
    \includegraphics[width=\textwidth]{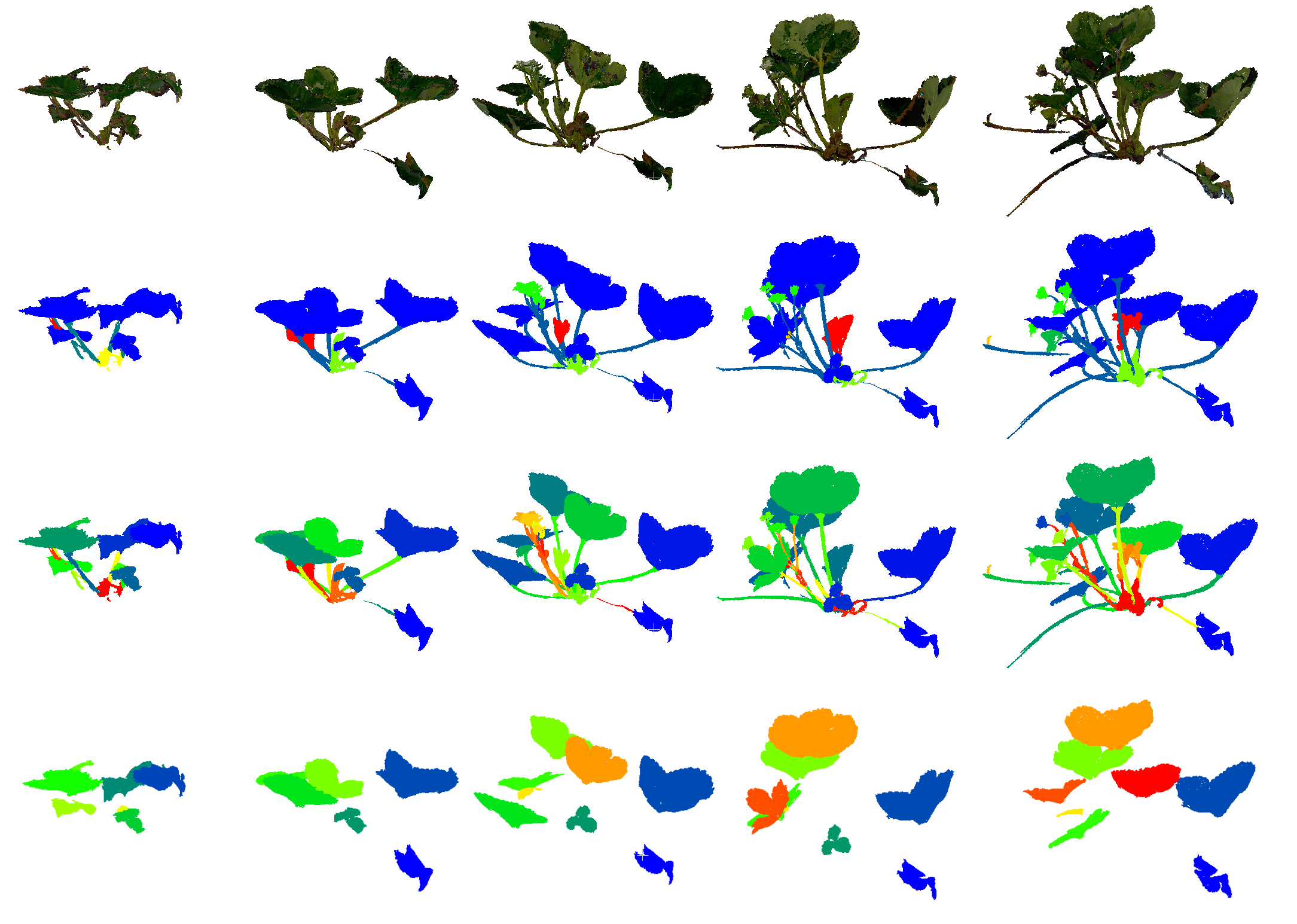} 
    \caption{Sample time-series showing a single plant (A2) over the first five time-steps, displaying the colour point clouds (row 1), semantic segmentation (row 2), instance segmentation (row 3) and temporally consistent leaf instance annotations (row 4). Note that the background and scanning table classes have been filtered out.}
    \label{fig:time-series}
\end{figure}

Of the full number of point clouds, 13 are annotated with class and instance labels - the process for which is described in Sec.~\ref{sec:annotation}. Additionally, the first five scans of plant A2 and the first two scans of B1 are supplied with temporally consistent leaf instance labels, as a demonstration of how this can be achieved for tracking purposes. The annotation of the dataset is summarised in Table \ref{tab:coverage} and a sample of a time series from the dataset is demonstrated in Fig.~\ref{fig:time-series}, to highlight semantic and instance annotations.

\begin{table}[hbt!]
    \centering
    \scriptsize
    \begin{tabular}{|c|c|cccccccccccccc|}
    \hline
         &Time step & 1 & 2 & 3 & 4 & 5 & 6 & 7 & 8 & 9 & 10 & 11 & 12 & 13 & 14 \\ \hline         
         \multirow{6}{*}{\rotatebox[origin=c]{90}{Plant number}} &A1&$\lgwhtcircle$ &  $\lgwhtcircle$ &  $\lgwhtcircle$ &  $\lgwhtcircle$ &  $\lgwhtcircle$ &  $\lgwhtcircle$ &  $\lgwhtcircle$ &  $\lgwhtcircle$ &  $\lgwhtcircle$ &  $\lgwhtcircle$ &  $\lgwhtcircle$ &  $\lgwhtcircle$ &  $\lgwhtcircle$ &  $\lgwhtcircle$ \\
         
         &A2&\textcolor{burgundy}{$\lgblkcircle$} &  \textcolor{burgundy}{$\lgblkcircle$} &  \textcolor{burgundy}{$\lgblkcircle$} &  \textcolor{burgundy}{$\lgblkcircle$} &  \textcolor{burgundy}{$\lgblkcircle$} &  $\lgwhtcircle$ &  $\lgwhtcircle$ &  $\lgwhtcircle$ &  $\lgwhtcircle$ &  $\lgwhtcircle$ &  $\lgwhtcircle$ &  $\lgwhtcircle$ &  $\lgblkcircle$ &  $\lgblkcircle$ \\
         
         &A3&$\lgwhtcircle$ &  $\lgwhtcircle$ &  $\lgwhtcircle$ &  $\lgwhtcircle$ &  $\lgwhtcircle$ &  $\lgwhtcircle$ &  $\lgwhtcircle$ &  $\lgwhtcircle$ &  $\lgwhtcircle$ &  $\lgwhtcircle$ &  $\lgwhtcircle$ &  $\lgwhtcircle$ &  $\lgwhtcircle$ &  $\lgwhtcircle$ \\
         
         &B1&\textcolor{burgundy}{$\lgblkcircle$} &  \textcolor{burgundy}{$\lgblkcircle$} &  $\lgwhtcircle$ &  $\lgwhtcircle$ &  $\lgwhtcircle$ &  $\lgwhtcircle$ &  $\lgwhtcircle$ &  $\lgwhtcircle$ &  $\lgwhtcircle$ &  $\lgwhtcircle$ &  $\lgblkcircle$ &  $\lgblkcircle$ &  $\lgblkcircle$ &  $\lgblkcircle$ \\

        & B2&$\lgwhtcircle$ &  $\lgwhtcircle$ &  $\lgwhtcircle$ &  $\lgwhtcircle$ &  $\lgwhtcircle$ &  $\lgwhtcircle$ &  $\lgwhtcircle$ &  $\lgwhtcircle$ &  $\lgwhtcircle$ &  $\lgwhtcircle$ &  $\lgwhtcircle$ &  $\lgwhtcircle$ &  $\lgwhtcircle$ &  $\lgwhtcircle$ \\

         &B3&$\lgwhtcircle$ &  $\lgwhtcircle$ &  $\lgwhtcircle$ &  $\lgwhtcircle$ &  $\lgwhtcircle$ &  $\lgwhtcircle$ &  $\lgwhtcircle$ &  $\lgwhtcircle$ &  $\lgwhtcircle$ &  $\lgwhtcircle$ &  $\lgwhtcircle$ &  $\lgwhtcircle$ &  $\lgwhtcircle$ &  $\lgwhtcircle$ \\
    \hline              
    \end{tabular}
    \caption{Dataset coverage in terms of annotation. $\lgwhtcircle$ = only point clouds available, $\lgblkcircle$ = instance annotations available, as well as ground truth skeletons of stems and \textcolor{burgundy}{$\lgblkcircle$} = temporally consistent leaf instance annotations available. }
    \label{tab:coverage}
\end{table}

\subsection{Collection procedure}
\begin{figure}
    \centering
    \begin{subfigure}[t]{0.45\textwidth}
    \centering
    \includegraphics[height=5.5cm]{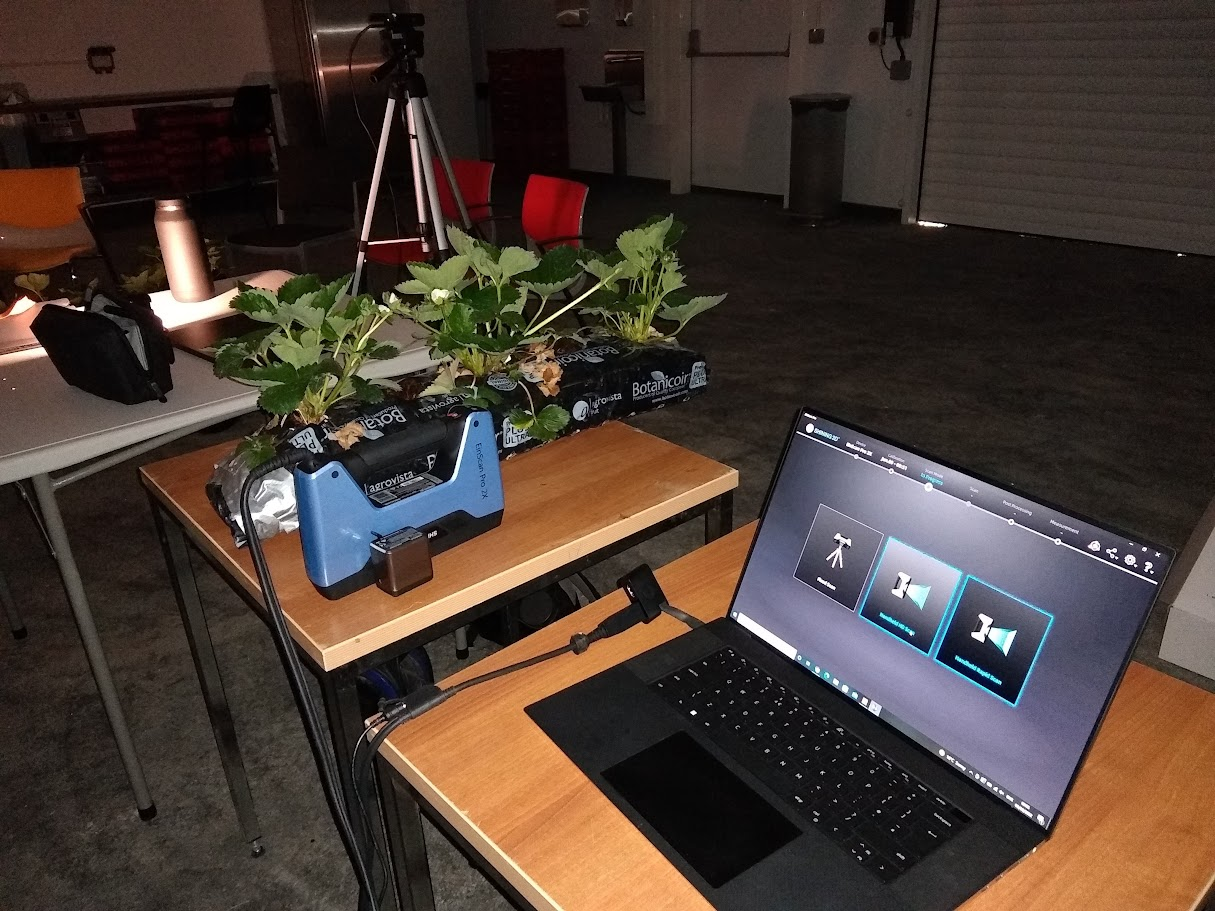}       
    \end{subfigure}%
   \begin{subfigure}[t]{0.45\textwidth} 
   \centering
  \includegraphics[height=5.5cm]{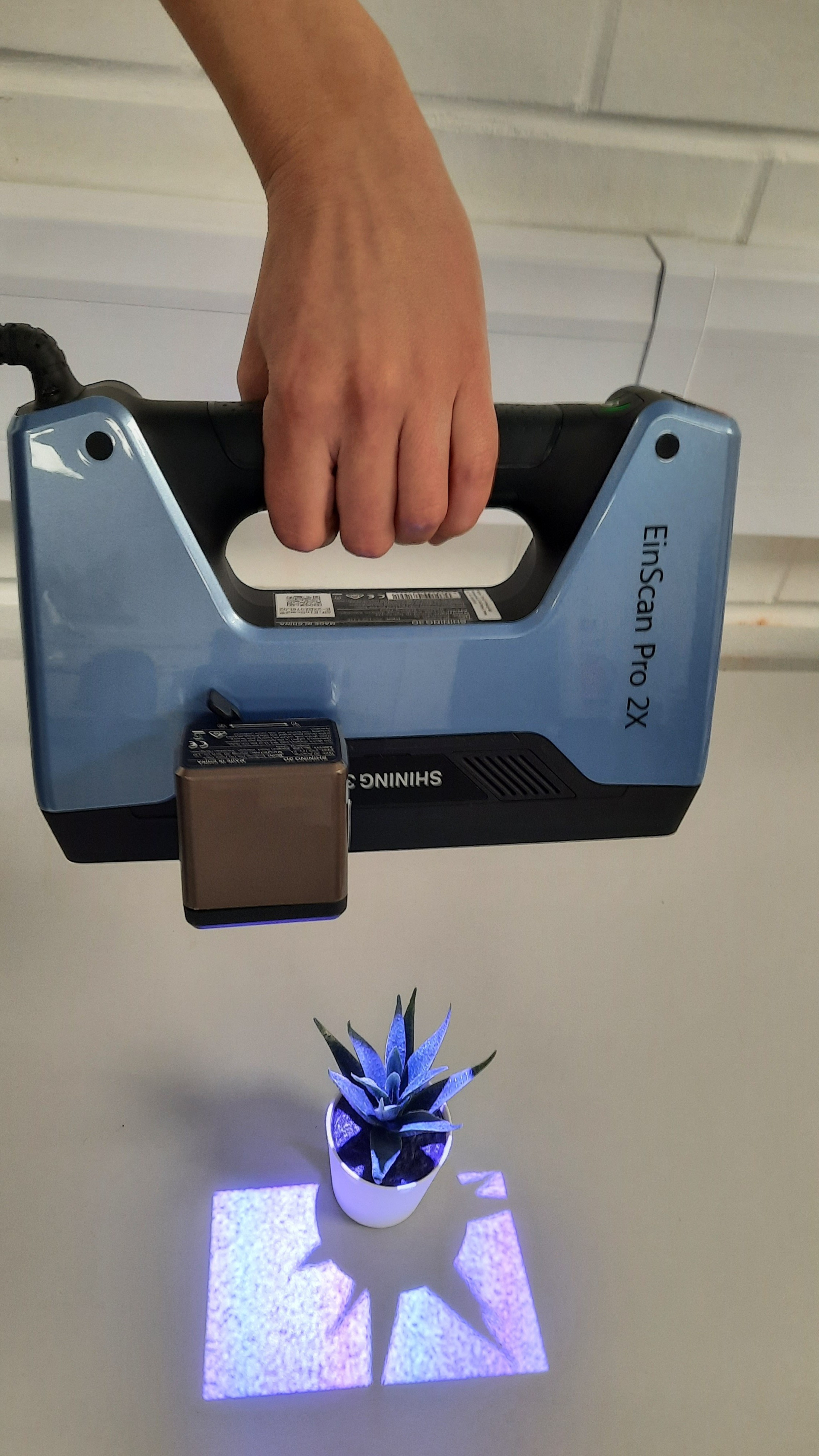}
    \end{subfigure}
    \caption{Setup of the EinScan Pro 2X Plus operated in `rapid handheld' mode. Plastic plant and lighting were used for demonstration purposes only.}   
    \label{fig:setup}
\end{figure}

The dataset was collected in indoor, controlled conditions with low light. The plants were scanned at least once weekly using an EinScan Pro 2X Plus Scanner\footnote{\url{https://www.einscan.com/handheld-3d-scanner/2x-plus/}}, which operates through a structured light projection and produces high-fidelity point clouds and meshes. The scanning setup is depicted in Fig.~\ref{fig:setup}. This scanner was used in `Rapid Handheld Scan' mode, alignment was automatically performed on extracted features and the colour pack was used, providing colour information for each 3D point. The scanner was calibrated before each use to achieve the expected volumetric accuracy of 0.1 mm. Each scan took upwards of 30 minutes to complete to best capture the scanning candidate and to ensure there were as few holes in the data as possible.

Each sequence of scans of the same plant was subsequently aligned manually into a common frame using CloudCompare\footnote{https://www.cloudcompare.org/main.html}, however, a common global coordinate frame between different plants can not be assumed. 

\subsection{Annotation} \label{sec:annotation}
\begin{figure}
    \centering
    \includegraphics[width=\textwidth]{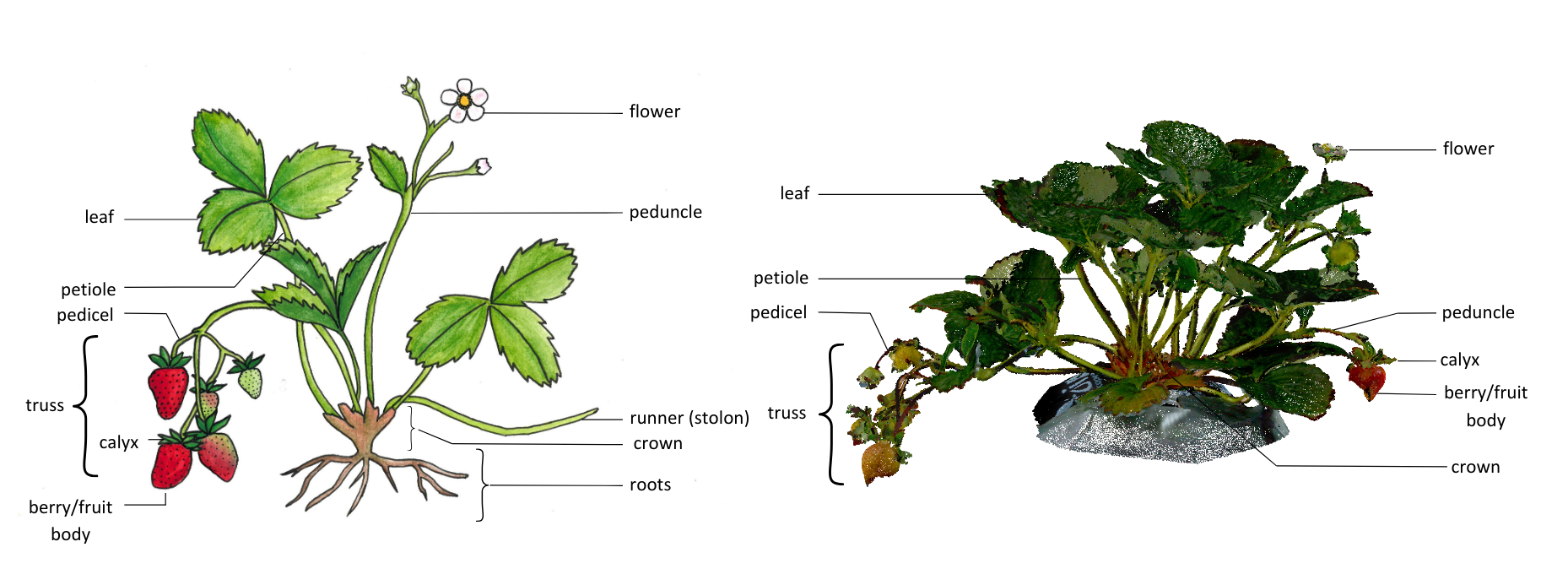}
    \caption{Anatomy of a strawberry plant. Left: annotated illustration. Right: annotated sample from the LAST-Straw dataset.}
    \label{fig:anatomy}
\end{figure}

For 13 of the 84 scans, we provide two sets of point-wise annotations: i) class labels and ii) instance labels.
We use seven classes to distinguish different plant organs and two additional classes for background information inadvertently captured during the scanning process. 
The semantic classes are 1) leaf or leaflet, 2) stem, including petiole, peduncle, pedicel, and stolon, 3) berry, 4) flower, 5) crown, 6) background, 7) other plant part, 8) scanning table, and 9) emergent leaf or leaflet.
A terminology reference for the anatomy of a strawberry plant is provided in Fig.~\ref{fig:anatomy}.

All annotations were performed manually using the Segments.ai\footnote{\url{Segments.ai}} 3D annotation tool by an annotator and edited by a separate reviewer. Importantly, the exact separation of the plants into functional organs and instances, i.e. densely labelling point clouds into disjoint regions, is not trivial and somewhat subjective. To minimise ambiguity, we started the labelling process by focusing on the regions that offer the clearest distinctions and proceeded to more uncertain areas. In case of diverging assessment from the annotator and reviewer, we additionally consulted HD video footage of the plant taken at the same time as the scan and employed the following rules for the categorisation of ambiguous areas:
an emergent leaf or leaflet has to be identifiable as a growing leaf but has not yet unfurled, yet visibly diverged from the axis of its petiole. We further define the cut-off point between flowers and berries based on whichever occurs first: either the protruding cone shape of the berry becomes distinctly visible at the centre of the structure, or a noticeable colour change is observed in the receptacle, transitioning from yellow to white. The crown class is defined as any central structure around the emergence point of the plant that is not otherwise identifiable.
Finally, the class `other plant part' consists of small structures associated with newly emerging organs that are not yet characterisable.

While not the primary object of the scan, we have decided not to crop out the background and scanning table classes. We intentionally retain these to serve as reference points, providing context on the capture setup and aiding in the manual alignment of scans. The background class includes soil, the grow bags housing the plants, parts of neighbouring plants visible in the scan, and any other objects captured in the background and can easily be filtered out where annotations are provided. No further pre-processing was performed. 

For tracking demonstration purposes (see Sec.~\ref{sec:tracking}) we additionally manually identified the temporally consistent identities of all leaf instances in the first five scans of plant A2 and the first two scans of plant B1 using CloudCompare. Only the instance labels for leaves in these seven scans can be considered temporally consistent. Please note that as a rule all annotations are nominal only and do not reflect any natural order such as the emergence date of an organ. 

\subsection{Skeleton annotation}

\begin{figure}[!ht]
    \centering
    \includegraphics[width = 0.75\textwidth]{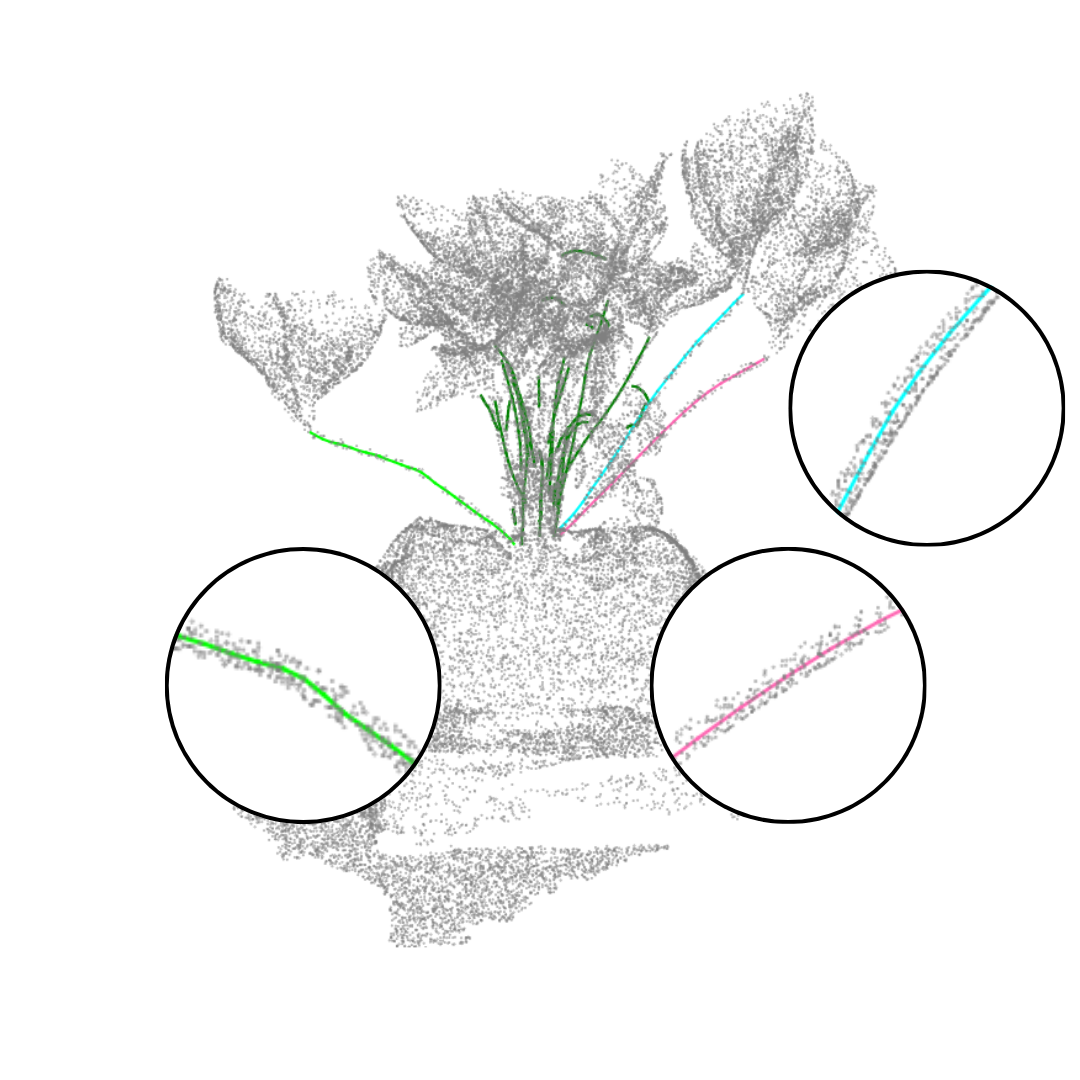}
    \caption{Ground truth skeletonisation of stem class superimposed on the original point cloud (downsampled for visualisation). Best viewed in colour.}
    \label{fig:gt_skel_all}
\end{figure}

From the 13 annotated samples, we extracted 267 instances of the stem class. A ground truth skeleton was generated for each stem using PlantScan3D \cite{plantscan3d}, which incorporates some existing skeletonisation algorithms and functions and provides an interface for placing and dragging skeleton points. These can then be exported as nodes and parsed to a more usable \textit{.ply} format. An initial skeletonisation result was obtained using the implementation of \citet{xu2007}. All excess branches were removed and the skeleton was manually corrected by dragging the skeleton points, smoothed and adjusted to give an approximation of the ground truth, through the centre of the stem. In cases where a partial point cloud had been captured, due to occlusion while scanning, a best estimate is used to place the ground truth skeleton to account for this. This mainly consisted of observing the curvature of the stem and approximating where the centre would be if not occluded. An example of all the skeletons for a particular plant superimposed on the original point cloud is provided in Fig.~\ref{fig:gt_skel_all}.

\section{Phenotyping} \label{sec:phenotyping}
The presented raw data alone offers the opportunity for visual examination of the complete 3D morphology at different stages in a plant's developmental process and side-by-side comparison of growth stages. In this section, we demonstrate the further processing steps allowing for the extraction of additional biologically relevant phenotypic measures.

\subsection{Segmentation}
A recent work by Harandi and colleagues~\cite{harandi2023make} offers a comprehensive overview of 3D plant segmentation methods. The overall consensus drawn is that, as of now, there is no standard approach that universally addresses plants in general or 3D data captured using different acquisition procedures. 
The majority of discussed methods actually require very little or no labelled training data but are usually highly customized towards a specific crop and require careful manual tuning.

Binary segmentation of plant versus background regions, for example, can often be accomplished through straightforward thresholding in a suitable colour space. In cases where a small amount of training data is available, a standard support vector machine (SVM) can be trained on fast point feature histograms (FPFH) of point clouds for classifying points into stem versus leaf categories (e.g.~\cite{magistri2020}). ~\cite{harandi2023make} compiles an overview of further approaches, often grounded in classical computer vision techniques translated into the 3D domain, such as region growing or clustering techniques, which can achieve dense semantic and instance segmentation with careful tuning. 

However, the authors specifically highlight Machine Learning (ML) algorithms as an immensely promising avenue for exploration. While many well-studied ML methods have shown great success for segmentation in other applications, translating these achievements to the unstructured point cloud input is proving challenging. Additionally, the high architectural complexity and considerable variation in thickness and density across different regions of the same plant further contribute to the difficulty of the task. The adaptation and customisation of ML models for plant phenotyping have lagged behind broader advancements, primarily due to the necessity for large and diverse labelled datasets. In ML, data availability is crucial, especially when aiming to generalise, such as across different plant species.

More recently there has been considerable progress in this area:
In Pheno4D \cite{schunck2021}, the authors demonstrate successful semantic and instance segmentation using three suitable network architectures (PointNet, PointNet++, LatticeNet). Additionally, it has been shown that training on simulated data \cite{heiwolt2021deep} as well as employing data augmentation strategies \cite{xin20233d} can yield significant benefits when annotated data is scarce.

Successful 3D segmentation is commonly the first step enabling an array of advanced phenotyping methods.
Binary segmentation into plant matter and background offers the ability to estimate the plant volume. In \cite{neumann2015dissecting}, Neumann and colleagues compute a `digital volume' measure from pixel counts of multiple 2D side-view and top-down images of plants. The study finds digital volume to correlate highly with manual measurements of biomass, serving as a non-destructive proxy measurement for plant biomass. 
With the full 3D model being available in our data, we filter out the background, discretise each plant point cloud into a voxel grid with a resolution of $1$~mm$^3$, and estimate plant volume as the total volume of occupied voxels. At the time of the first scan, plant A2 is substantially larger with an estimated volume of $85.95$~cm$^3$ than plant B1  at $56.64$~cm$^3$, but by the end of the measurement period, A2 is outgrown by B1 with $233.97$~cm$^3$ to $253.42$~cm$^3$. 



Semantic segmentation is needed to extract similar geometric measurements for each type of plant organ separately, and a combined semantic and instance segmentation allows for the extraction of phenotypes from individual organs. However, further steps have to be taken to extract these phenotypes. 
For example, a mesh needs to be constructed to extract leaf area, and skeletonisation must be performed to find the medial axis of stem-like structures.

\subsection{Leaf surface reconstruction}
To estimate leaf area, a surface has to be reconstructed from the leaf point cloud, which is a non-trivial task in itself. 
In \cite{ando_robust_2021} the authors weigh the advantages and disadvantages of existing model-based and model-free methods and introduce a novel approach which reduces distortions by digitally unfurling leaf point clouds before producing a mesh representation. \cite{boukhana2022geometric} further offers a comparative study on the performance of existing leaf area estimation methods and explores the criteria affecting the robustness of each studied method in detail. However, the examined methods largely target unifoliate leaves and make a number of assumptions about their typical shape and relative dimensions which are not compatible with the trifoliate leaves of strawberry plants.
In this work, we use three classical 3D triangle surface meshing approaches, which do not rely on explicit shape assumptions as a simple baseline. 

\paragraph{2.5D Delaunay triangulation}
Using the Leaf Axis Determination procedure from~\cite{ando_robust_2021}, we first find the local right-hand coordinate frame for each leaf, so that the $x$ and $y$ axes point in the two perpendicular directions covering the leaf's largest variance in 3D space. The point cloud is then projected into that two-dimensional plane, which maintains the largest spatial spread of points. Ignoring the z-axis, the 2D coordinates are then triangulated and the resulting mesh structure is applied back to the full 3D coordinate set. In this work we use the implementation provided by CloudCompare \cite{CloudCompare}.

\paragraph{Ball-pivoting algorithm}
The Ball-Pivoting algorithm \cite{bernardini1999ball} produces a triangle mesh gradually by rolling a ball of a fixed radius across the point cloud. A new surface triangle is created between three points whenever the ball can rest on all three points without containing any other point within its radius. For this work we specifically use the MeshLab \cite{meshlab} implementation, which offers an auto-guess for the ball radius based on the point cloud density.  

\paragraph{Zabawa method}
Finally, we also use the surface area estimation procedure introduced by Zabawa and colleagues in \cite{zabawa2021automated}. Specifically devised for the use on plant leaves, this pipeline consists of statistical outlier removal, uniform sub-sampling, and mesh generation via three passes of the Ball-Pivoting algorithm where the radius each time is scaled dependent on the point cloud's average distance between points, and an operation to close holes in the resulting mesh. 

\paragraph{\textit{in silico} ground truth}
A ground truth mesh was constructed for the first five instances of leaf 6 of plant A2 manually using Blender~\cite{blender}. Consulting the colour point cloud and HD video footage, we manually edited the mesh created by the Ball-Pivoting algorithm. Typical edits included closing superfluous holes in the mesh (e.g. see the region marked by a small red circle in Fig.~\ref{fig:meshing_artefacts}~d and removing obvious artefacts of the scanning and meshing process.
A frequent artefact observed in 3D point clouds of plants, for example, occurs in planar sections of the leaf. With a small volume yet large surface area leaves pose a unique challenge to 3D sensing. Coordinates captured from perspectives on the top and bottom surface of the leaf are not always perfectly planar, but instead may exhibit a slight offset perpendicular to the section plane. Some surface reconstruction methods produce two overlapping parallel surface sections with an offset between them in the affected regions. For our ground truth meshes, we aimed to remove any such duplicate surfaces. See the two large circled regions in Fig.~\ref{fig:meshing_artefacts}~d for examples.

\subsubsection{Evaluation}
To compare the three automated meshing approaches, we report the mean absolute percentage error (MAPE) between the estimated ($A$) and ground truth ($A^{*}$) surface area across $n=5$ leaves:
\begin{equation}
    \textrm{MAPE} = \frac{1}{n} \sum_{i=1}^{n} \left|\frac{A^{*}_i - A_i}{A^{*}_i}\right|.
\end{equation}


\begin{table}[]
    \centering
    \begin{tabular}{|l|l|lll}
    \cline{1-2}
    Method                      & MAPE  &  &  &  \\ \cline{1-2}
    2.5D Delaunay Triangulation~\cite{CloudCompare} & 2.620 &  &  &  \\
    Ball-Pivoting Algorithm~\cite{bernardini1999ball} & 0.262 &  &  &  \\
    Zabawa Method~\cite{zabawa2021automated}               & 0.165 &  &  &  \\ \cline{1-2}
    \end{tabular}
    \caption{Mean absolute percentage error between estimated and ground truth surface area across five leaves for three triangle meshing approaches.}
    \label{tab:leaf_area_metrics}
\end{table}

\begin{figure}
    \begin{subfigure}[t]{\textwidth}
            \centering
        \includegraphics[width=\textwidth]{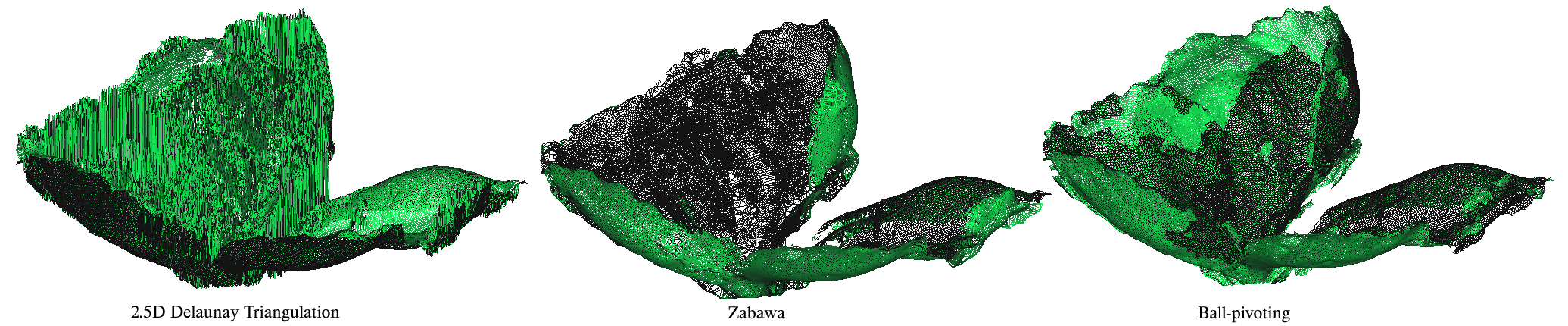}
        \subcaption[]{Wiremesh rendering of triangle meshes reconstructed from the point cloud of plant A2, leaf 6, at time step 2 using 2.5D Delaunay triangulation (left), the Zabawa method (centre), and the Ball-Pivoting algorithm (right).}
    \hspace{-0.25cm}
    \end{subfigure}
    \begin{subfigure}[t]{0.42\textwidth}
            \centering
        \includegraphics[width=\textwidth]{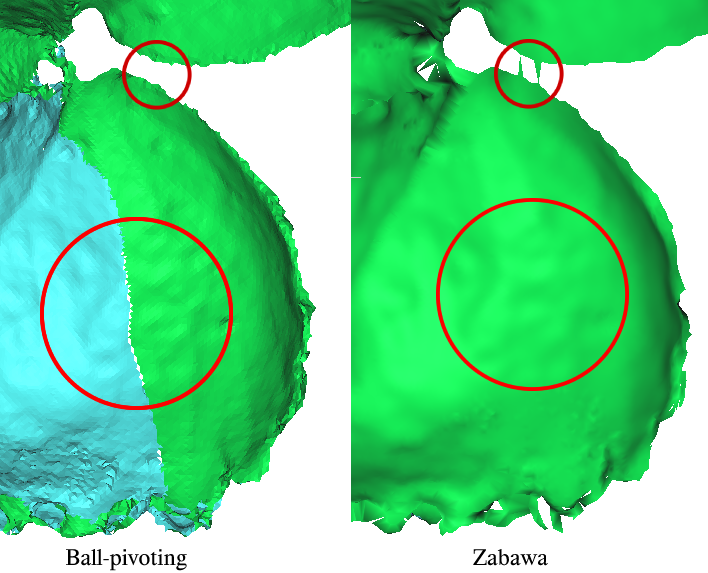}
        \subcaption[]{Close-up of triangle meshes taken from plant A2, leaf 6, scan 3, with colour representing triangle surface normals. Red circles highlight differences in the structure characteristic of the Ball-Pivoting algorithm (left) and the Zabawa method (right).}
    \end{subfigure}
    \hspace{0.25cm}
    \begin{subfigure}[t]{0.58\textwidth}
            \centering
        \includegraphics[width=\textwidth, trim={2cm 0 2cm 0 },clip]{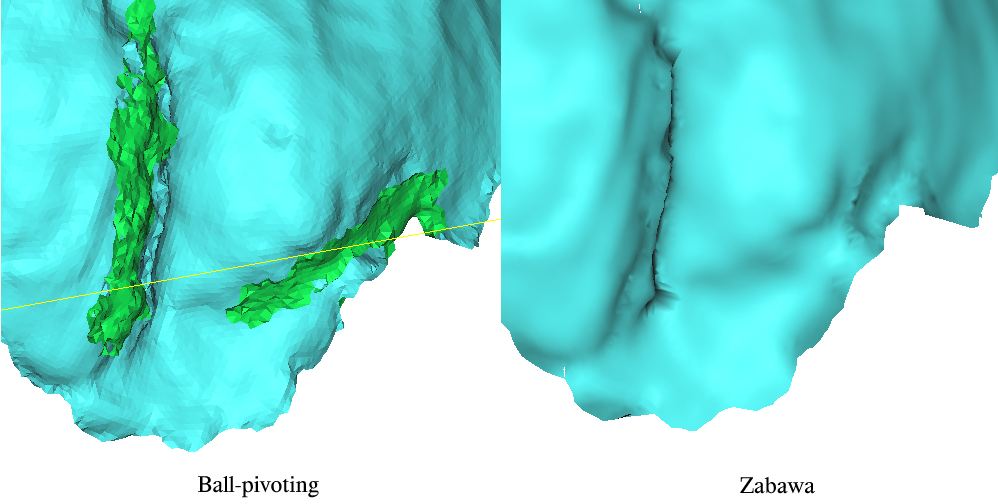}
        \subcaption[]{Close-up of triangle meshes taken from plant A2, leaf 6, scan 3, with colour representing triangle surface normals. Green areas in the left panel indicate regions where two separate parallel surfaces were reconstructed (Ball-Pivoting algorithm), where in the right panel points were instead merged into a single surface (Zabawa method).}
    \end{subfigure}
    \begin{subfigure}[t]{\textwidth}
            \centering
        \includegraphics[width=.9\textwidth]{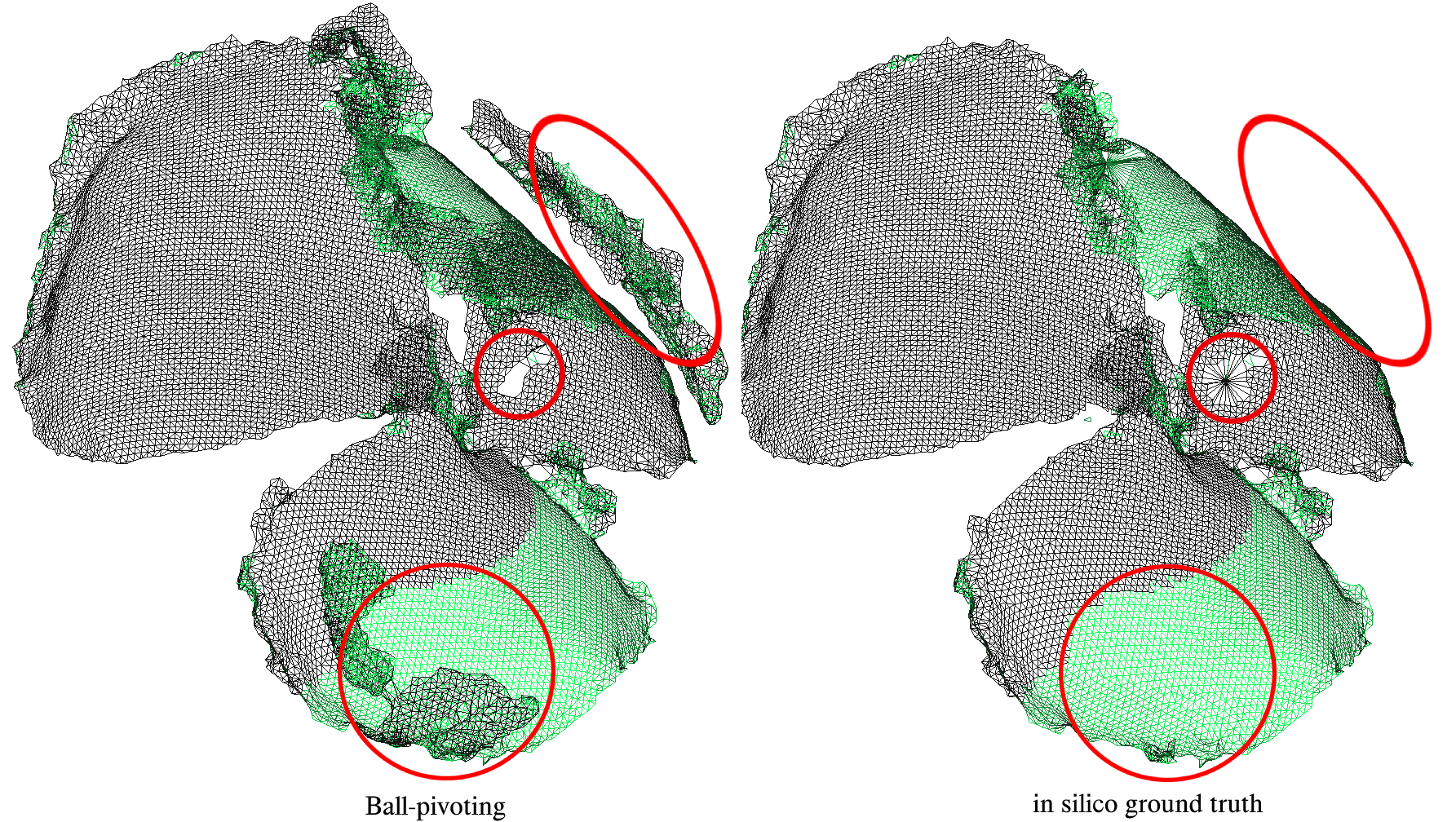}
        \subcaption[]{Wiremesh rendering of triangle meshes reconstructed from the point cloud of plant A2, leaf 6, at time step 1 using the Ball-Pivoting algorithm (left), and manually edited \textit{in silico} ground truth (right).}
    \end{subfigure}
        \caption{Visual comparison of automated mesh reconstructions (a-c) and an example of how the \textit{in silico} ground truth was created (d).}
        \label{fig:meshing_artefacts}
\end{figure}

Qualitatively, the three presented methods exhibit characteristic strengths and weaknesses. 
This Delaunay algorithm typically works best for flat structures, struggles to accurately capture steep sloping surfaces, and fails to produce flat surfaces where there is any furling or overlap in the z-direction. Fig.~\ref{fig:meshing_artefacts}~a shows one example where the bend in the leaf causes multiple surface sections to overlap in the z-dimension. The 2.5D Delaunay Triangulation procedure then erroneously interconnects vertices vertically, purely based on their proximity in the x-y-plane. The result is a spiking effect which greatly overestimates the true surface area.

The success of the Ball-Pivoting algorithm largely depends on an even point sampling density. Thus, the additional pre- and post-processing steps added to the algorithm in the Zabawa method, which largely affects the point density, visibly affect the resulting interconnectivity of regions that stay otherwise separated in the Ball-Pivoting algorithm. Fig. \ref{fig:meshing_artefacts}~b and c highlight examples of this. In both cases, the Zabawa mesh appears more smoothed and less detailed along the edges. In panel b, the large circle indicates areas where the Zabawa method succeeds in closing holes and constructing more consistently oriented, continuous surfaces. At the same time, the approach displays a tendency to connect surfaces across larger gaps, where not desirable. Panel c further highlights some robustness of the Zabawa method to the artefact of erroneous parallel surface sections discussed above.  

Through qualitative and quantitative assessment, the Zabawa method emerges as the closest match to the ground truth. However, it is essential to note that this small sample limits the feasibility of hypothesis testing. Further exploration is warranted to assess the method's suitability for leaf area estimation on trifoliate, 
heavily furled, and irregularly shaped leaves.

\subsection{Skeletonisation} \label{sec:skeletonisation}

Skeletonisation of plant point cloud is a process which produces a set of skeleton points connected by edges to form a graph, reducing the point cloud to a single thin representative structure. Plant skeletons have been utilised for organ tracking \cite{chebrolu2020, magistri2020} and the production of these is an essential step for interpreting the topology and geometry of stem-like structures in plants and are necessary for the computation of phenotypes such as length. Semantic information is a step which can substantially improve the skeletonisaton process, through the extraction of skeleton points from individual organs \cite{magistri2020}. In this section, we exploit the knowledge of individual instances of the stem class from the annotations to provide skeletonisation benchmarks on our dataset. 

\subsubsection{Quantitative assessment of skeleton}
To determine the quality of the skeleton for use in phenotyping tasks, namely the measurement of length, we introduce methodology to quantitatively assess an estimated skeleton against a ground truth skeleton. Let each skeleton $S$ consist of a set of vertices $V$, connected by a set of edges $E$, forming a non-directed graph. Given two skeletons, the groundtruth $S_g$ and the estimate $S_e$, we need to assess the agreement in topology and geometry between them. 

The number of vertices in each estimated skeleton depends strongly on the algorithm used and can vary from the order of 10 to 1000 (in skeletons from L1-medial skeletonisation). Computed metrics need to be comparable across such skeletons of different vertex densities. To address the difference in vertex density between skeletons for assessment, a dense graph can be created, as in \citet{horn2021}, splitting edges to obtain a maximum edge length $s_{dense}$. We do not remove nodes to re-sample the skeleton uniformly with spacing $s_{dense}$ between the nodes, as would have the potential to alter the geometry of the skeleton.

\paragraph{Polyline graph matching}
To determine which nodes in $S_e$ match those in $S-g$, we perform polyline graph matching to determine the number of true positives (TP), false positives (FP) and false negatives (FN). The TP count is the number of nodes in $S_e$ that match those in $S_g$, FP is number of nodes in $S_e$ that do not match nodes in $S_g$, nor fall on an intermediate line segment between nodes, and FN is the number of nodes in $S_g$ that are not matched in $S_e$.

To compare two polyline graphs in 3D, consisting of nodes and edges, we compute a cost matrix with rows corresponding to $S_g$ and columns to $S_e$. For each pair of points, we set the cost equal to the Euclidean distance between them if the distance is less than or equal to $t_{match}$ and otherwise to an arbitrarily selected high value (1000). The Hungarian algorithm \cite{crouse2016} is then used to find the minimal cost of assignment. 

We then create a dictionary $D$ to indicate the matches between the two graphs. In the case where there are unmatched vertices in $S_g$, all values in the cost matrix will be equal, and we update $D$ to reflect this by changing the value to -1. $D$ thus yields a 1-1 matching at the vertex density level of $S_g$.  

\begin{figure}
    \centering
    \includegraphics[width=0.7\textwidth]{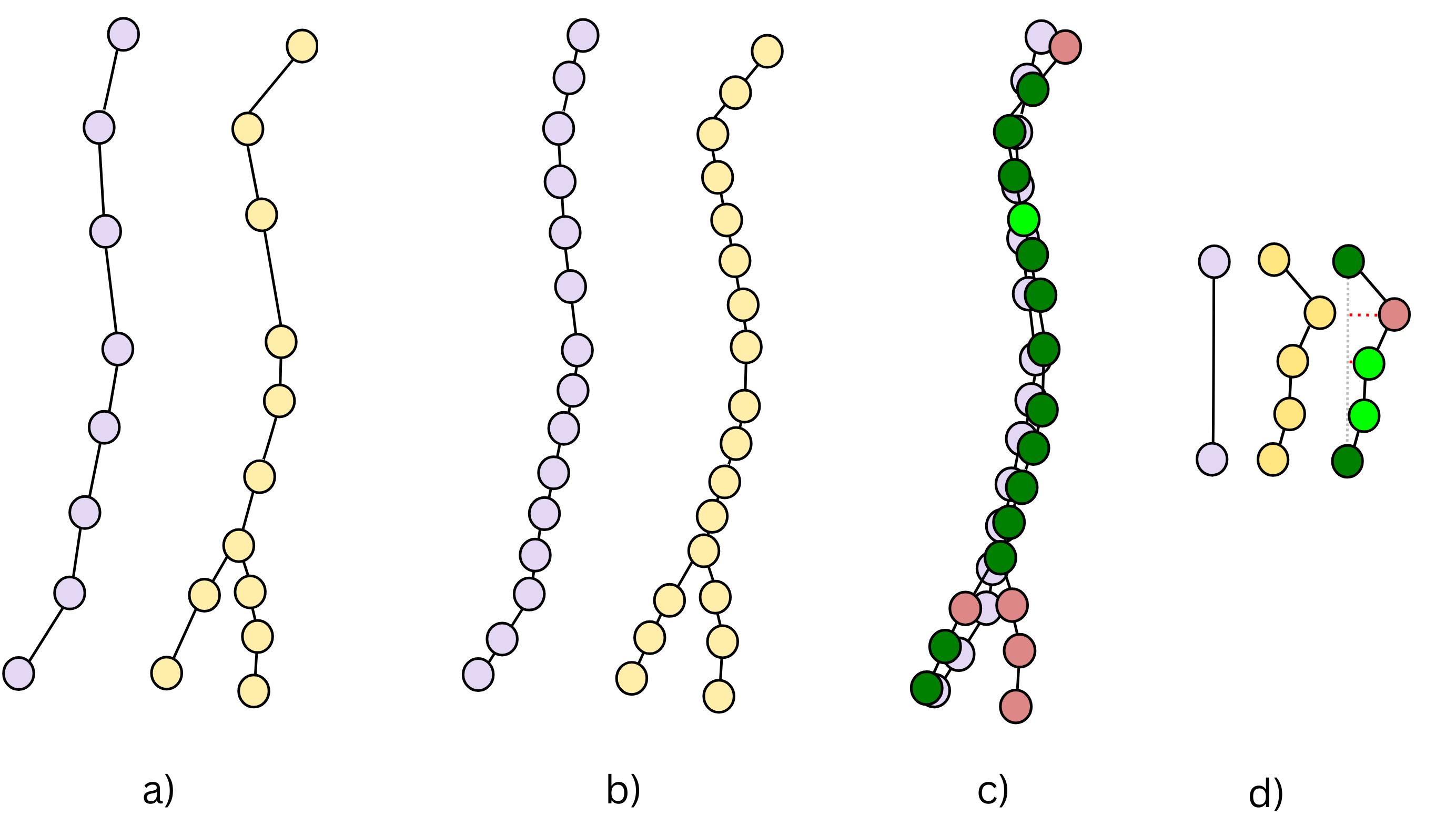}
    \caption{The polyline matching process between ground truth (purple) and estimate (yellow) graphs. With a) raw graphs b) dense graphs and c) matched graphs. In d) we show an illustrative example of the case where $S_e$ is denser than $S_g$ or nodes in $S_e$ fall between two matched nodes - nodes which fall within a threshold $t_{line}$ but which are not assigned using the Hungarian algorithm are marked as positive (light green) but are not included in the true positive count, to keep the number of true positives consistent with the ground truth.}
    \label{fig:polyline_matching}
\end{figure}

If the graph is denser than the maximum length $s_{dense}$ of the dense graph, then extra unmatched vertices that fall between TP would be marked falsely as unmatched (FP). To avoid this, we split the ground truth graph into individual branches (including disconnected components), then compute the line segments for all the nodes within the branch and, finally, consider the vertex in $S_e$ to be matched if it falls within a threshold $t_{line}$ away from the line. FP are then only those nodes which are still unmatched after this procedure. This is illustrated in Fig.~\ref{fig:polyline_matching}. 

Given that the rough diameter of a stem in a strawberry plant is 3.17 mm wide, we set $t_{match}$ and $t_{line}$ to an eighth of this (0.396). The smallest computationally feasible dense graph requires that $s_{dense}$ be set to 0.3. Following the creation of the dense graph, we then apply polyline graph matching to allow the calculation of precision, recall and F1-score, defined as follows:

\begin{equation}
    \textrm{Precision} = \frac{\textrm{TP}}{\textrm{TP}+\textrm{FP}},\quad \textrm{Recall} = \frac{\textrm{TP}}{\textrm{TP}+\textrm{FN}}, \quad \textrm{F1-score} = 2\frac{\textrm{Precision} \cdot \textrm{Recall}}{\textrm{Precision} + \textrm{Recall}}.
\end{equation}

and to calculate additional metrics relating to length and topology. High precision implies that there are very few deviations from the ground truth skeleton points beyond the threshold, while high recall implies good coverage of these. Deviations can be due to incorrect topology (branching differently) or misalignment of the skeleton within the points (falling outside the threshold distance). 

To determine if the skeleton has a good topological match, we additionally report on the number of branch end points ($N_{end}$). This is because for the skeletonisation of single stems, we theoretically expect that there will only be two branch end points, and higher numbers indicate excess branching. The connectivity of the skeleton is also important, so we report on the number of fragmented segments ($N_{seg}$) in the computed skeleton. Finally, as we ultimately wish to assess the quality of the skeleton for phenotyping, we report the fraction of the total matched length ($L_{matched}$). In the case of branches being predicted in an application or study, a single end-to-end path would need to be selected to measure length, without the presence of a ground truth skeleton to select the correct path. We therefore additionally report on the MAPE between the longest path $L$ and ground truth length $L^{*}$, as this gives a measure of the likely error that can be expected in real-world measurements
\begin{equation}
\textrm{MAPE} = \frac{1}{n} \sum_{i=1}^{n} \left|\frac{L^{*}_i - L_i}{L^{*}_i}\right|,
\end{equation}
where n is the number of stems.

\subsubsection{Measurement of length}
In this section, we compare three skeletonisation methods against ground truths to quantify their performance for the task of extracting the phenotype of stem length. The first method calculates a single-source shortest path through a graph constructed from local neighbourhoods of points and uses a manually placed root node \cite{xu2007}. This algorithm was initially developed for the skeletonisation of trees, but has been used in studies using other plants (eg. \cite{chaudhury2020, chebrolu2020}). The second algorithm, L1-medial skeletonisation \cite{huang2013}, is a general-purpose skeletonisation algorithm but has been used in plant studies \cite{chaudhury2020, chebrolu2020}. The final method makes use of self-organising maps (SOMs) and has been found to provide more robust skeletonisation for plant organ tracking in maize and tomato \cite{magistri2020}. Importantly, all three of these methods have open-source implementations available. We refer to these methods as `Shortest Path', L1-medial and SOM going forwards.

\begin{figure}[!htbp]
  \centering    
  \includegraphics[width=\textwidth]{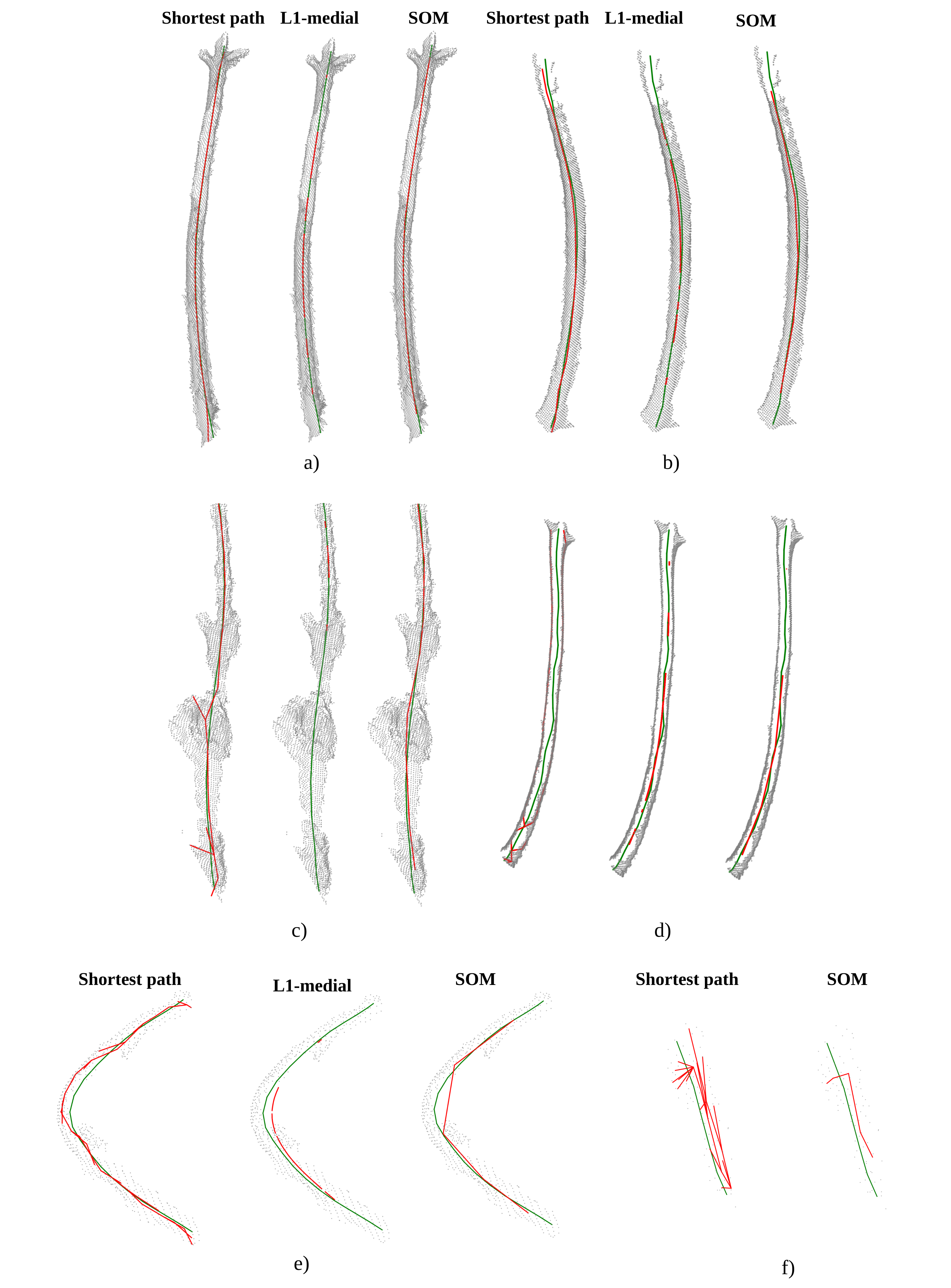}
  \caption{Skeletonisation results (red) compared to ground truth (green). Please note that this is image is a single view of a 3D point cloud and corresponding skeleton. Best viewed in colour.}
  \label{fig:skel_results}
\end{figure}

To compute skeletons using the Shortest Path method, a root node was manually placed at the bottom of the stem and a bin count of 10 was used. The initial radius used in L1-medial skeletonisation was set to 3.17, which is similar to the diameter of a stem. The number of nodes for the SOM approach was adjusted in the same way as the authors in \citet{magistri2020}, based on a fraction of the number of points present. A sample of skeletonisation results is shown in Fig.~\ref{fig:skel_results} and metrics computed across all 267 stem samples are reported in Table \ref{tab:skeletonisation_metrics}. It should be noted that the L1-medial medial method failed to produce results for 16 samples, so these were excluded from the mean metrics reported. 
 
\begin{table}[!hbtp]   
    \centering
    \small
    
    \begin{subtable}{\textwidth}
    \centering
    \begin{tabular}{|c|c|c|c|c|c|c|c|}
    \hline
         Method & Precision & Recall & F1   \\  \hline
         Shortest Path~\cite{xu2007} & 0.72 $\pm$ 0.23 &\textbf{ 0.70 $\pm$ 0.22 }&\textbf{ 0.69 $\pm$ 0.26} \\
         L1 medial~\cite{huang2013} & 0.71 $\pm$ 0.38 & 0.31 $\pm$ 0.21 & 0.38 $\pm$ 0.33\\ 
         SOM~\cite{magistri2020} & \textbf{0.74 $\pm$ 0.28} & 0.53 $\pm$ 0.22 & 0.60 $\pm$ 0.29 \\    
     \hline
    \end{tabular}
    \end{subtable}
    
    \vspace{1em} 
    
    \begin{subtable}{\textwidth}
    \centering
    \begin{tabular}{|c|c|c|c|c|c|c|c|}
    \hline
         Method & $L_{matched}$ ($\%$) & MAPE & $N_{end}$ & $N_{seg}$  \\  \hline
         Shortest Path & \textbf{ 0.74 $\pm$ 0.36 }& \textbf{0.17 $\pm$ 0.72} & 5.34 $\pm$ 6.78 & \textbf{ 1.00 $\pm$ 0.00}\\
         L1 medial &  0.27 $\pm$ 0.22 & 0.74 $\pm$ 0.24 & 8.79 $\pm$ 4.88 & 4.40 $\pm$ 2.44\\ 
         SOM & 0.53 $\pm$ 0.44 & 0.24 $\pm$ 0.44 & \textbf{2.12 $\pm$ 0.44} & 1.04 $\pm$ 0.20\\ 
     \hline
    \end{tabular}
    \end{subtable}
   
    \caption{Mean values and standard deviation of skeletonisation assessment compared to the ground truth. }
    \label{tab:skeletonisation_metrics}
\end{table}

From the reported metrics we are able to interpret the success of the various methods in predicting suitable skeleton points, interrogate the topology of the skeleton in terms of branching (the number of end points, theoretically = 2) and the connectivity of the skeleton (the number of segments, theoretically = 1). Furthermore, we are able to interpret how well methods perform for the ultimate goal of skeletonisation in this case - extracting the length of the stem. 

From the metrics in Table \ref{tab:skeletonisation_metrics}, we immediately observe a relatively high level of variance, indicating a non-normal distribution in the dataset in terms of the difficulty posed by samples to the skeletonisation methods. Fig.~\ref{fig:skel_results} reflects some of the variations within the dataset, illustrating different challenges that can be presented, including noise, occlusion (including changing levels of occlusion along the length of the stem) and low point density or numbers.

All methods report a similar level of precision from dense skeleton matching but differ substantially in terms of recall. Here we observe that the L1-medial method does not handle this dataset well, with a very low recall due to a highly fragmented skeleton (averaging at 4.4 segments per stem) with poor coverage reflected in recall, length match and a high MAPE in length. This is visually notable in Fig.~\ref{fig:skel_results}. This method furthermore fails to produce a skeleton for 16 samples when the number of points is low ($<$ ~300). This indicates that although it is reported to give good results on general objects, this method is not suited to extracting biologically relevant length measurements from a dataset such as this.

The performance and behaviour of the remaining two methods differ in that the Shortest Path method produces fully connected skeletons which have many branches, while the SOM method tends to produce far fewer branches - observable both from the number of end points and also particularly notable Fig.~\ref{fig:skel_results}~{c-f}. The SOM method, however, produces skeletons that are more truncated towards the end points - resulting in a loss of coverage and thus a reduction in recall and length match scores, see Fig.~\ref{fig:skel_results}~{a, b, d}. The Shortest Path method thus produces a far more robust recall score than SOMs and has a higher fraction of matched length. 

Challenges in the dataset do, however, affect both methods. Samples $d$ and $e$ in Fig.~\ref{fig:skel_results} demonstrate behaviour in the case of samples where occlusion is present. In $d$ Shortest Path does not recognise the stem as a single stem, but rather as having two branches, while the SOM method produces only partial coverage. Sample $e$ illustrates a case where the level of occlusion changes as the stem curves, resulting in skeletons that either move towards the existing points (Shortest Path) or cut off the curve (SOM). This illustrates that further research into skeletonisation methods that can handle partial occlusion of points would be beneficial for phenotyping studies. 
 
The mean MAPE that could be expected in a phenotyping operation indicates that an error of 17\% could be achieved by selecting the longest path if applied to data with a similar distribution of difficult and easy samples as in the assessed dataset. However, this varies considerably depending on the sample and associated challenges. 

From these results, we would suggest that the Shortest Path method is a fair purpose skeletonisation algorithm for plant structures, with reasonable performance for the purpose of extracting a meaningful length measurement from individual stems. This algorithm has the drawback that a high level of branching is present, which poses the further challenge of requiring that a user must prune any excess branches on single stems. However, the SOM approach does provide the best match for topology with the least excess branching and low fragmentation, although the extracted length from this method does not indicate good reliability for measurements of length. None of the assessed skeletonisation methods handle difficult samples well, such as those containing occlusion, resulting in off-centre and non-biologically relevant skeletons. We thus motivate that further research into handling skeletonisation for real-world phenotyping is needed.

\section{Tracking} \label{sec:tracking}
As highlighted in \cite{das2019leveraging}, plants grow rapidly and non-uniformly. Beyond interpreting each observation as a static snapshot in time, the ability to represent dynamic changes in plant morphology and measure novel temporal phenotypes promises to improve breeders' and researchers' understanding of plant development, genome-environment interactions, and early indicators for plant quality or environmental stresses. 

\subsection{Tracking composition over time}
\begin{figure}[!ht]
    \centering
    \includegraphics[width=\textwidth]{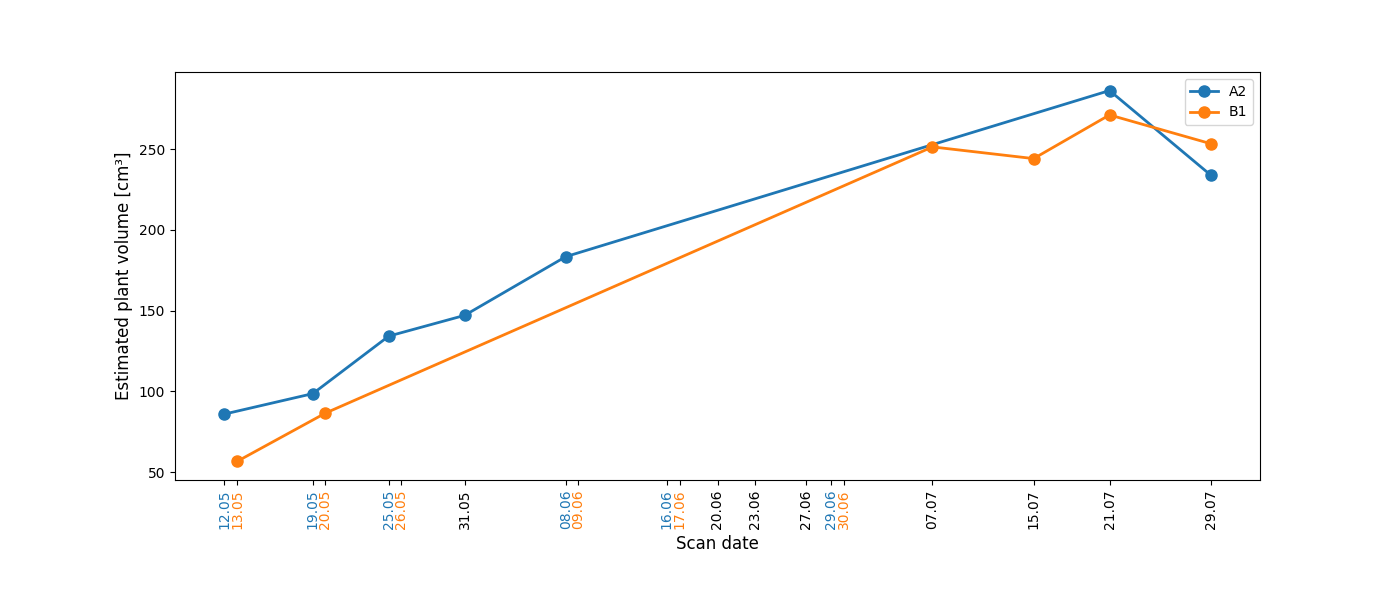}
    \caption{Estimated plant volume over time for all 13 annotated scans in the dataset. For context on the time scale, all scan dates are listed, regardless of whether annotations and thus volume estimates are available. Dates printed in black indicate that scans of both plants were recorded that day. Best viewed in colour.}
    \label{fig:volume_over_time}
\end{figure}

Given access to multiple scans of the same plant, any measurement taken of the plant as a whole can be plotted against the time scale to reveal new insight into the temporal dimension of plant development. In Fig.~\ref{fig:volume_over_time} we contextualise the estimated plant volume of each annotated scan within the series of repeated scans of the same plant across the data collection period - together characterising the overall plant growth. 

\begin{figure}[!ht]
    \centering
    \begin{subfigure}[t]{0.51\textwidth}
        \centering
        \includegraphics[width=\textwidth, trim={0.9cm 0 1cm 0 },clip]{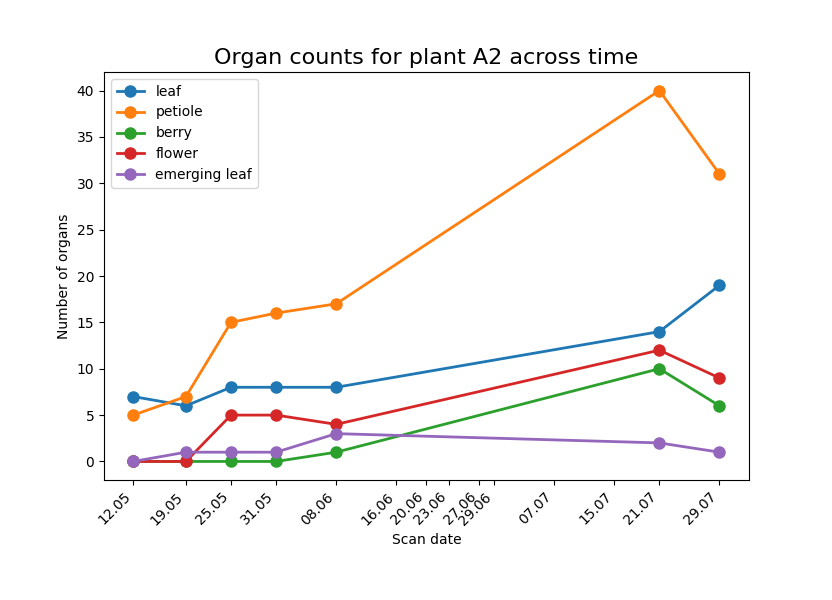}
    \end{subfigure}
    \begin{subfigure}[t]{0.48\textwidth}
        \centering
        \includegraphics[width=\textwidth, trim={0.9cm 0 1cm 0 },clip]{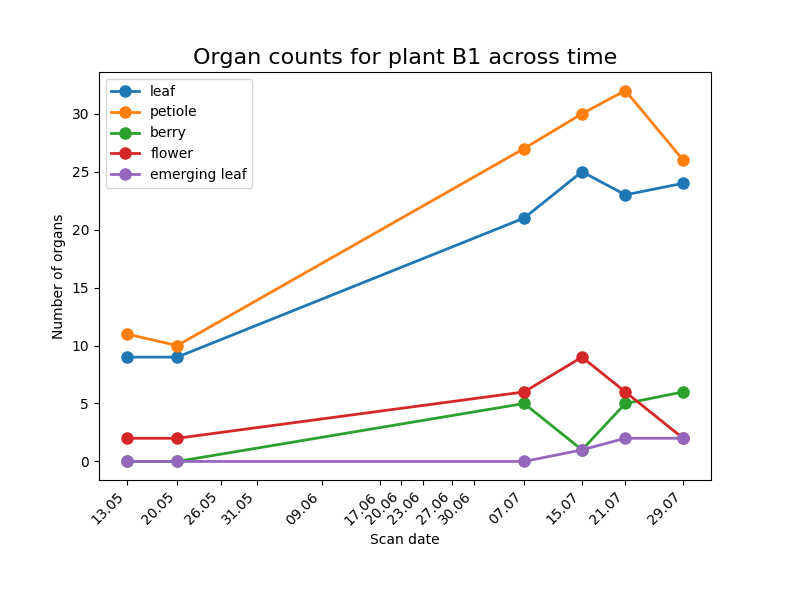}       
    \end{subfigure}
    \caption{Instance counts of each organ for plants A2 and B1 over time, where annotations are available. For context on the time scale, all scan dates are listed, regardless of whether annotations and thus organ counts are available.}
    \label{fig:organ_counts}
\end{figure}

In Fig.~\ref{fig:organ_counts} we track the instance counts per semantic class over time. Both time series individually reveal discernible growth spurts and highlight significant events, such as the transformation of flowers into berries. Together they aid in contrasting the two plants' compositions at similar stages in their respective life-cycles. 

Next, to measure temporal phenotypes of individual plant components, it is necessary to find correspondences between consecutive scans. The data association process, involving the identification of individual organs and linking these organs to a unique temporally consistent identity, is non-trivial. Existing approaches are typically specific to the type of organ, thus in the following sections we discuss how tracking of individual leaves and petioles across repeated scans might be accomplished.

\subsection{Leaf instance tracking}
Existing approaches associate and track leaves across time by their similarities in their global location in 3D space \cite{magistri2020}, by their relative topological location within the graph-like branching structure of the plant \cite{chebrolu2021registration, pan2021multi}, or by characteristic shape features of individual leaves \cite{Heiwolt2023Statistical}.

For demonstration purposes, we have manually annotated this association for the first five scans of plant A2 and the first two scans of plant B1, and assigned temporally consistent instance labels to leaves in those scans.
Knowing the individual temporally consistent identity of an organ now allows us to track and observe trends in organ-specific changes, such as the surface area of single leaves. 

\begin{figure}[!ht]
    \centering
    \begin{subfigure}[t]{\textwidth}
        \centering
        \includegraphics[width=.8\textwidth]{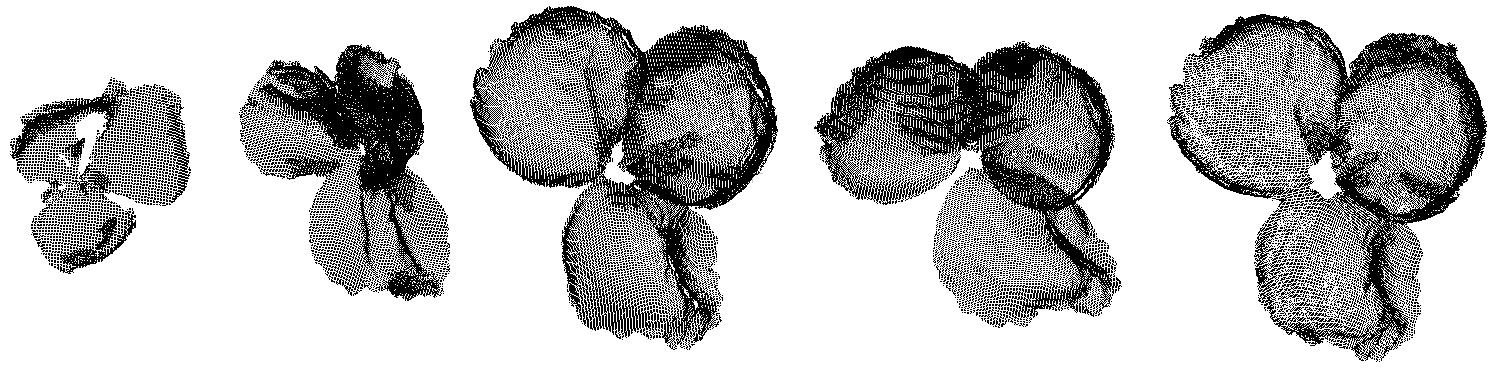}
    \end{subfigure}
        \begin{subfigure}[t]{\textwidth}
        \centering
        \includegraphics[width=.8\textwidth]{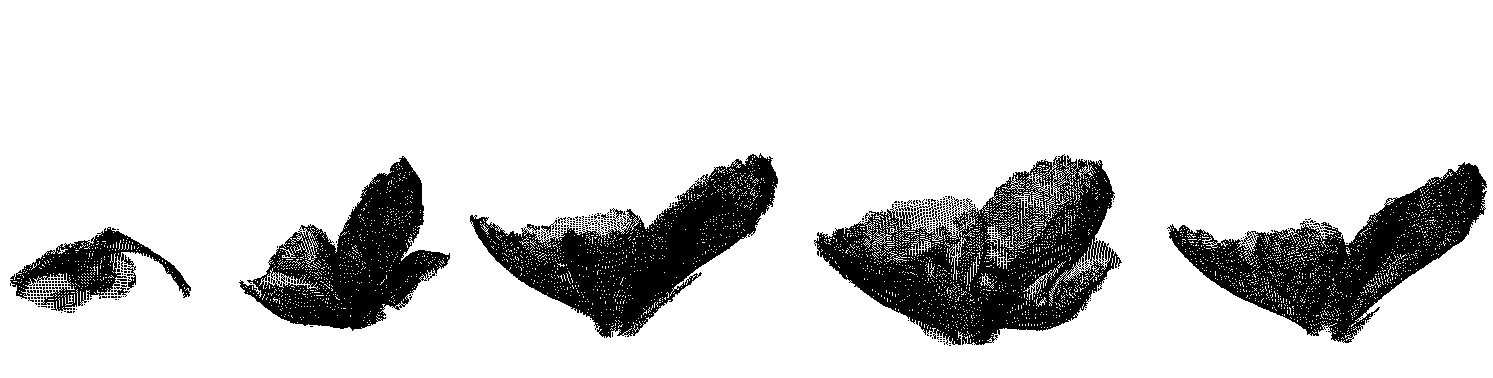}
    \end{subfigure}
    \begin{subfigure}[t]{\textwidth}
        \centering
        \includegraphics[width=\textwidth]{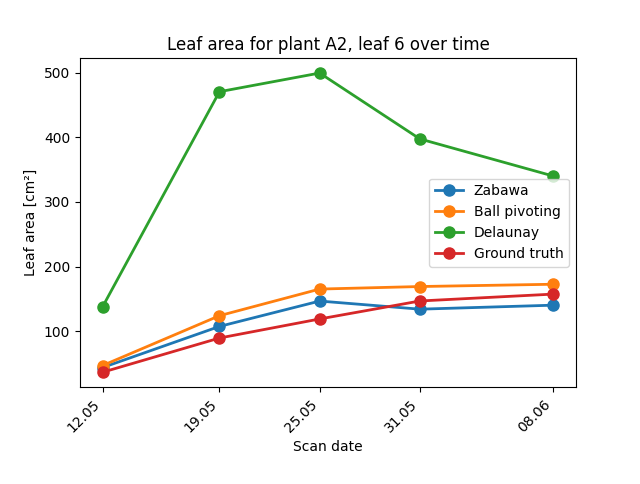}       
    \end{subfigure}
    \caption{The top shows a visualisation of the point clouds for leaf 6 of plant A2 over the first 5 scans, both from a top-down and side view. Bottom panel: Plot of the leaf's surface area across time as estimated from three different triangle-meshing procedures and the \textit{in silico} ground truth mesh. Best viewed in colour.}
    \label{fig:individual_leaf_area}
\end{figure}

\begin{figure}[!ht]
    \centering
    \begin{subfigure}[t]{0.49\textwidth}
        \centering
        \includegraphics[width=\textwidth]{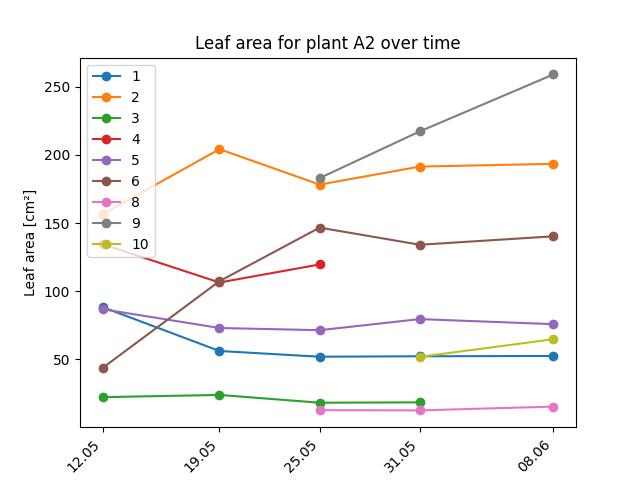}
    \end{subfigure}
    \begin{subfigure}[t]{0.49\textwidth}
        \centering
        \includegraphics[width=\textwidth]{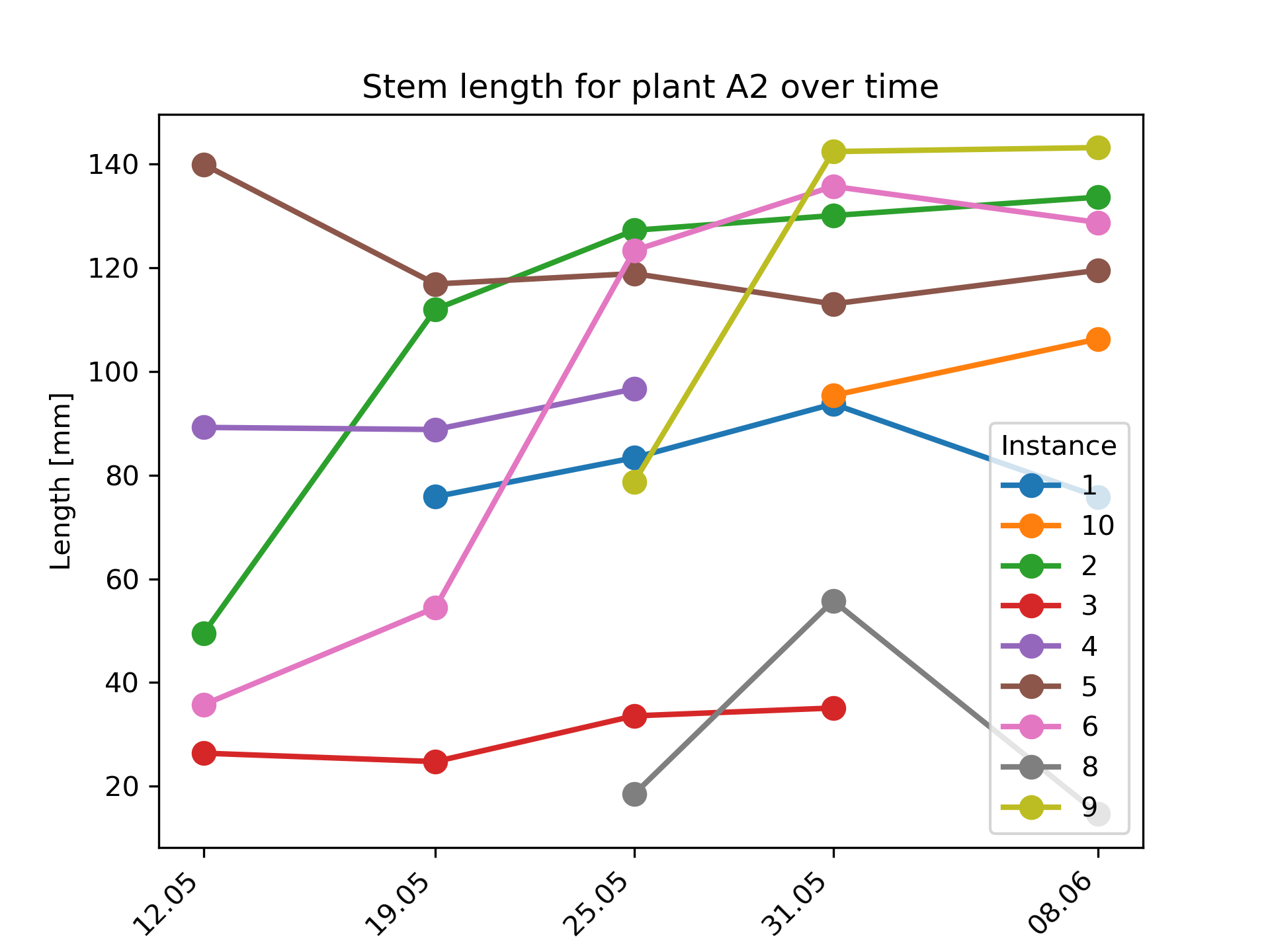}       
    \end{subfigure}
    \caption{Tracking of phenotypes over time for the first five time-steps for plant A2. (Left) Leaf area estimates using the Zabawa method. (Right) Petiole length estimates using the Shortest Path method. }
    \label{fig:measure_over_time}
\end{figure}

Fig.~\ref{fig:individual_leaf_area} features a visualisation of a single leaf as captured in the first five consecutive scans. Along with visualising its growth process, we also track its estimated leaf area across time. The observable growth in the rendered point clouds is clearly reflected in the corresponding area measurements. In Fig.~\ref{fig:measure_over_time}, we extend the plot to include surface area estimates for all leaf instances present within the same time span, measured using the Zabawa method. Instances where the estimated surface area appears to decrease with time can be explained largely by scanning and meshing artefacts, such as duplicate surfaces and holes in the scan. Future work on leaf area estimation remains challenging and will need to address the distinction between occlusions or missing data and holes in the leaves. The visual representation may appear chaotic, suggesting the non-uniformity of growth and underscoring the persistent challenges in area estimation.

\subsection{Petiole instance tracking} 
If we assume that leaf instance ids are global, it is then possible to track individual petioles semi-automatically. For each stem, we identify the end points of its skeleton and filter these, removing the end point closest to the crown, to obtain a set of potential leaf junction points. We reduce each leaf to a single point, the mean of the leaf points, and then apply the Hungarian algorithm to find the least cost of assignment in the distances between these two sets. By associating the petiole with a global leaf instance, we are then able to assign a global stem instance identifier to the petiole such that it is temporally consistent and track these over time, see Fig.~\ref{fig:measure_over_time}. 


\begin{figure}[!ht]
    \centering
    \includegraphics[width=.8\textwidth]{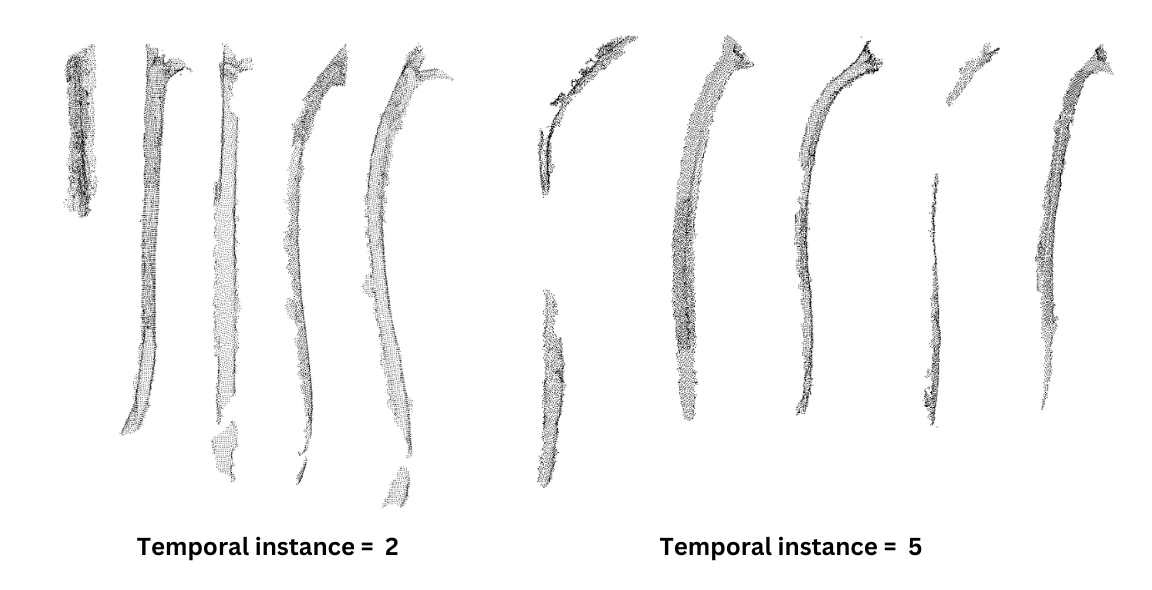}
    \caption{Point clouds associated with petiole instances 2 and 5 respectively. It is clear that for instance 2 there is an increase in length over time, while 5 illustrates how the challenges to phenotyping can cause a disruption to measurements.}
    \label{fig:inspect_petioles}
\end{figure}

Challenges to real-world phenotyping remain in that it is only possible to measure using the observable data. In cases when stems are sufficiently visible, we see the expected increase in stem length over time (for example, instance 2). However, in cases such as instances 1, 5 and 8, we observe in Fig.~\ref{fig:measure_over_time} that the length of the petiole decreases over time. Although this does not reflect the reality of the situation, in these cases the point clouds representing these stems are truncated or broken due to occlusion. The difference between an unoccluded (instance 2) and occluded (instance 5) stem can be seen in Fig.~\ref{fig:inspect_petioles}. Occlusion thus remains a major challenge to phenotyping and work towards addressing this is needed. Further challenges remain regarding tracking stem instances over time, particularly in tracking stems which are not petioles, so are not associated with global leaf ids. We highlight this as a challenge for future work.

\section{Conclusions and Future Work}
We contribute a dataset of high-resolution strawberry point clouds to assist in the development of spatio-temporal phenotyping tools. Through our demonstration of the phenotyping pipeline, we show how it is possible to use semantic segmentation to track plant volume and instance segmentation to track the emergence of particular organs and phenotypes extracted thereof, such as leaf area. We further highlight how applying skeletonisation methods allows the interpretation of length from stem-like structures. Finally, we observe that through the re-identification of leaves, we can track both leaf area and petiole length over time. 

The progress made here demonstrates the enormous potential of automated plant phenotyping to assist in revolutionising the plant breeding process of complex perennial crops such as strawberries. Precise and accurate quantification of the size and position of vegetative and reproductive organs will enable selection decisions to be made with a far greater degree of accuracy and precision than is possible currently and will permit repeat observations within a small time frame, guiding breeders to genotypes showing particular promise and enabling a greater proportion of the positive genetic variation within a breeding population to be selected and exploited. 

The impact of this work, however, extends beyond phenotyping applications. The automated assessment of the plant architecture enables the development of selective harvesting robots which require precise knowledge about the individual plant components such as stem or crop for effectively directing the picking apparatus~\cite{xiong2023location}. There is also a growing interest in developing in-field crop monitoring systems which require precise per-plant information for tasks such as automated nutrient and water control in glasshouses or vertical farming systems (e.g.~\cite{ren2023mobile}). Such technologies would revolutionise agriculture and help address growing issues with labour shortages and the sustainability of food production. The precise geometrical information about plants over time is also critical for the development, parametrisation and validation of structural-functional crop growth models which are important for different aspects of crop science (e.g.~\cite{soualiou2021functional}). 

Challenges do, however, remain at all steps of this pipeline. The development of improved segmentation algorithms with the ability to differentiate various plant organs is a pressing requirement. Notably, many existing 3D segmentation techniques, such as SAM3D~\cite{yang2023sam3d}, were originally designed for large man-made objects featuring regular geometric patterns of planar surfaces. Translating these techniques to the intricate and complex structure of plants is likely not straightforward, while the availability of annotated training data for task-specific segmentation is extremely time-consuming, making this largely unviable (upwards of 4 hours for samples in our dataset). This highlights the need for further exploration of semi-automated and unsupervised approaches. 
  
Furthermore, the showcased phenotyping methods rely on the completeness and high spatial resolution of our data, limiting their current applicability in real-world scenarios. The sensor used in this study cannot easily be deployed in the field due to high sensitivity to light interference, and even under optimal sensing conditions, achieving full coverage from diverse viewpoints around a plant remains challenging. In the field, limited freedom of movement and the proximity of neighbouring plants lead to frequent (self-)occlusion. While advancements in sensing technology are promising, phenotyping techniques will also need to overcome these challenges. Current skeletonisation and surface reconstruction algorithms are strongly affected by the quality of the point clouds, even point clouds such as this which were collected in highly controlled conditions. This indicates a critical need for the development of sensors and methods to translate phenotyping beyond such conditions to the field. Furthermore, to enable high-throughput phenotyping, the manual steps in the pipeline would need to be fully automated, replacing hand-held scanning with a robot or gantry system to carry sensors and software to integrate the different steps of the pipeline.

There is a need for a growing corpus of high-quality, annotation-rich 3D crop scans. With the contribution of the LAST-Straw dataset and that of Pheno4D, there are now three crops with such datasets publicly available (strawberry, maize and tomato). Future work would see the addition of more crops, particularly the inclusion of data which captures a range of developmental stages. 

Within the LAST-Straw dataset, there is still room for improvement, particularly in terms of increasing the number of scans annotated. The dataset will be updated as these become available. Further benchmarking could also be done for additional phenotypes, or alternative methods compared against those we have already addressed. Further research is also needed to address the challenges associated with phenotyping with real-world data, particularly how to handle partial occlusion. Finally, while petiole tracking can be addressed through the association with the attached leaf, further research is needed for tracking other stem-like structures in the plant, such as trusses, and the individual fruit thereof.


\paragraph{Funding acknowledgement}
This work was supported by the Engineering and Physical Sciences
Research Council [EP/S023917/1] 
and the Collaborative Training Partnership (CTP) for Fruit Crop Research.

\paragraph{Data availability statement}
The LAST-Straw dataset and supplementary code can be accessed via \url{https://lcas.github.io/LAST-Straw/}. Supplementary code for graph matching can be accessed via \url{https://github.com/LCAS/GraphMatching3D}.




 \bibliographystyle{elsarticle-num-names} 
 \bibliography{laststraw2024}





\end{document}